\documentclass[final,3p,times,twocolumn]{elsarticle}
\usepackage[utf8]{inputenc}

\usepackage{natbib}

\usepackage{subfigure}
\usepackage{graphicx}
\usepackage{amsmath}
\usepackage{amsfonts}
\usepackage{xcolor}
\usepackage{caption}
\usepackage{mathtools}
\usepackage{bm}
\usepackage{soul}
\usepackage{booktabs}
\usepackage{hyperref}

\usepackage{algorithm}
\usepackage{algpseudocode}

\usepackage{titlesec}

\usepackage{makecell, multirow, threeparttable}

\newcommand{\para}[1]{\vspace{.05in}\noindent\textbf{#1}}

\journal{Neural Networks}

\title{Multi-task Heterogeneous Graph Learning on Electronic Health Records}
\begin{document}

\author[inst1]{Tsai Hor Chan}

\affiliation[inst1]{organization={Department of Statistics and Actuarial Science, The University of Hong Kong},%Department and Organization
            addressline={Pokfulam Road}, 
            city={Hong Kong SAR},
            % postcode={00000}, 
            % state={State One},
            country={China}
            }

\author[inst1]{Guosheng Yin}
\author[inst2]{Kyongtae Bae}
\author[inst1]{Lequan Yu \corref{cor1}}
\cortext[cor1]{Corresponding author. Email address lqyu@hku.hk}
\affiliation[inst2]{organization={Department of Diagnostic Radiology, The University of Hong Kong},%Department and Organization
            addressline={Pokfulam Road}, 
            city={Hong Kong SAR},
            % postcode={22222}, 
            country={China}}
\begin{abstract}
  Learning electronic health records (EHRs) has received emerging attention because of its capability to facilitate accurate medical diagnosis.
  Since the EHRs contain enriched information specifying complex interactions between entities, modeling EHRs with graphs is shown to be effective in practice.
  The EHRs, however, present a great degree of heterogeneity, sparsity, and complexity, which hamper the performance of most of the models applied to them.
    Moreover, existing approaches modeling EHRs often focus on learning the representations for a single task, overlooking the multi-task nature of EHR analysis problems and resulting in limited generalizability across different tasks. 
  In view of these limitations, we propose a novel framework for EHR modeling, namely MulT-EHR (\underline{Mul}ti-\underline{T}ask EHR), which leverages a heterogeneous graph to mine the complex relations and model the heterogeneity in the EHRs.
    To mitigate the large degree of noise, we introduce a denoising module based on the causal inference framework to adjust for severe confounding effects and reduce noise in the EHR data.
  Additionally, since our model adopts a single graph neural network for simultaneous multi-task prediction, we design a multi-task learning module to leverage the inter-task knowledge to regularize the training process.
  % We adopt a shared weihgt ... to address the multi-task nature of EHR and further design a MT learning module...
  %
  Extensive empirical studies on MIMIC-III and MIMIC-IV datasets validate that the proposed method consistently outperforms the state-of-the-art designs in four popular EHR analysis tasks --- drug recommendation, and predictions of the length of stay, mortality, and readmission.
  Thorough ablation studies demonstrate the robustness of our method upon variations to key components and hyperparameters.
\end{abstract}

\maketitle

%---------------------------------------------------------
\section{Introduction}
%---------------------------------------------------------

\begin{figure*}
    \centering
    \includegraphics[width=0.85\textwidth]{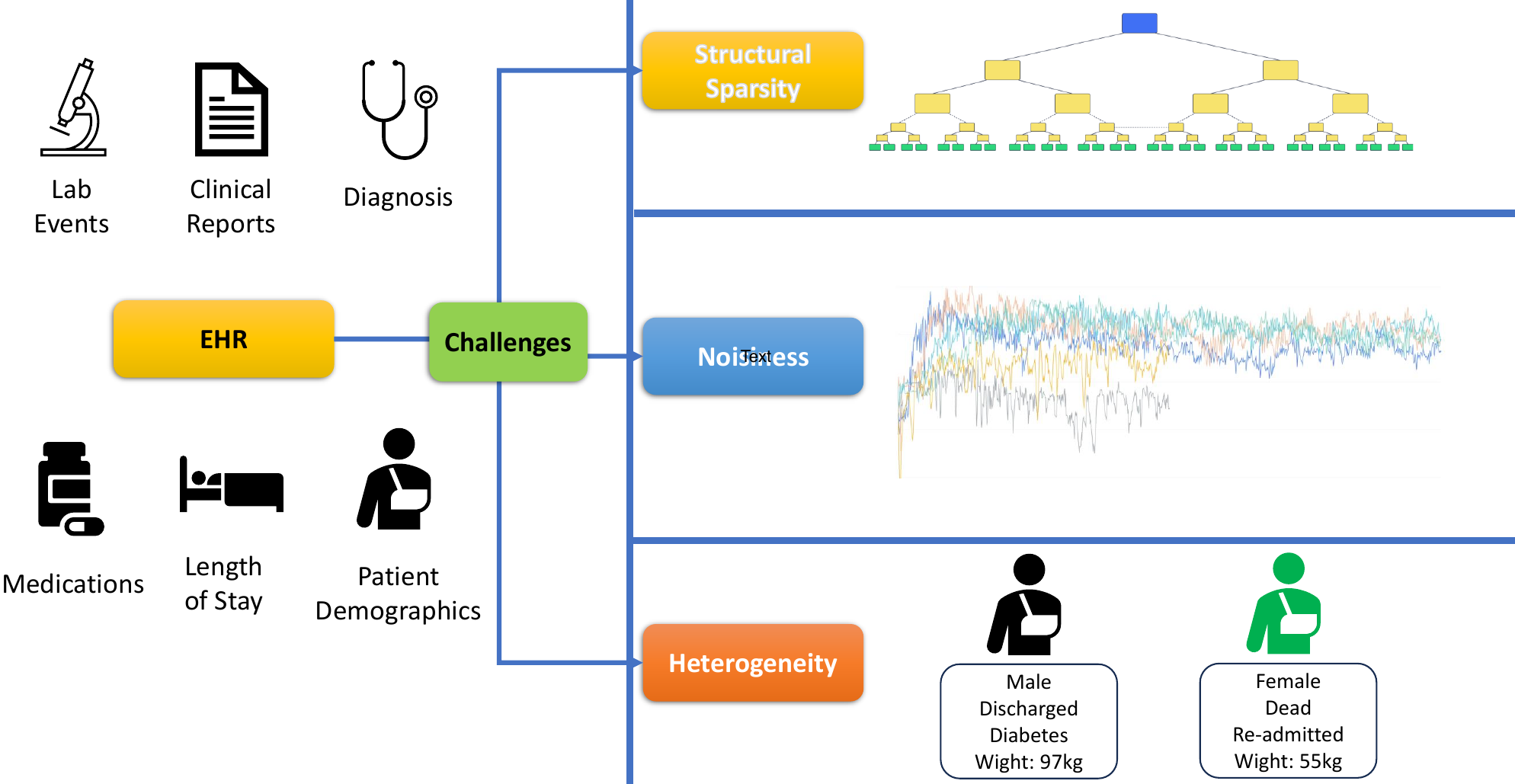}
    \caption{Illustration of EHR data and the challenges. EHRs contain enriched information on clinical visits of patients, including relevant medications and diagnoses. However, analysis of EHR data poses three challenges --- the sparsity in the data structure, noisiness, and heterogeneity of patients and their visits make it difficult to deliver accurate analysis.}
    \label{fig: challenges}
    \vspace{-0.3cm}
\end{figure*}

The process of clinical decision making heavily relies on the medical records of patients.
These records, however, present a great degree of heterogeneity, sparsity, and complexity in practice, making feature representation learning difficult.
Figure \ref{fig: challenges} illustrates the current challenges of EHR analysis.
In order to provide accurate clinical decisions, clinicians have to traverse through the complex records to search for evidential information, which is time-consuming and cost-ineffective.
%some tasks such as mortality prediction in the intensive care unit, the complexity of medical records makes clinicians difficult to provide an accurate diagnosis.
%
Recently, with the increasing availability of electronic health records (EHRs) in machine-readable forms, deep learning models have shown their powerful capability to mine the deep connections between medical entities and facilitate accurate decision making.
These deep learning models on EHRs have shown great potential in developing personalized medical treatment and improving healthcare quality.

\begin{figure}
    \centering
    \includegraphics[width=0.5\textwidth]{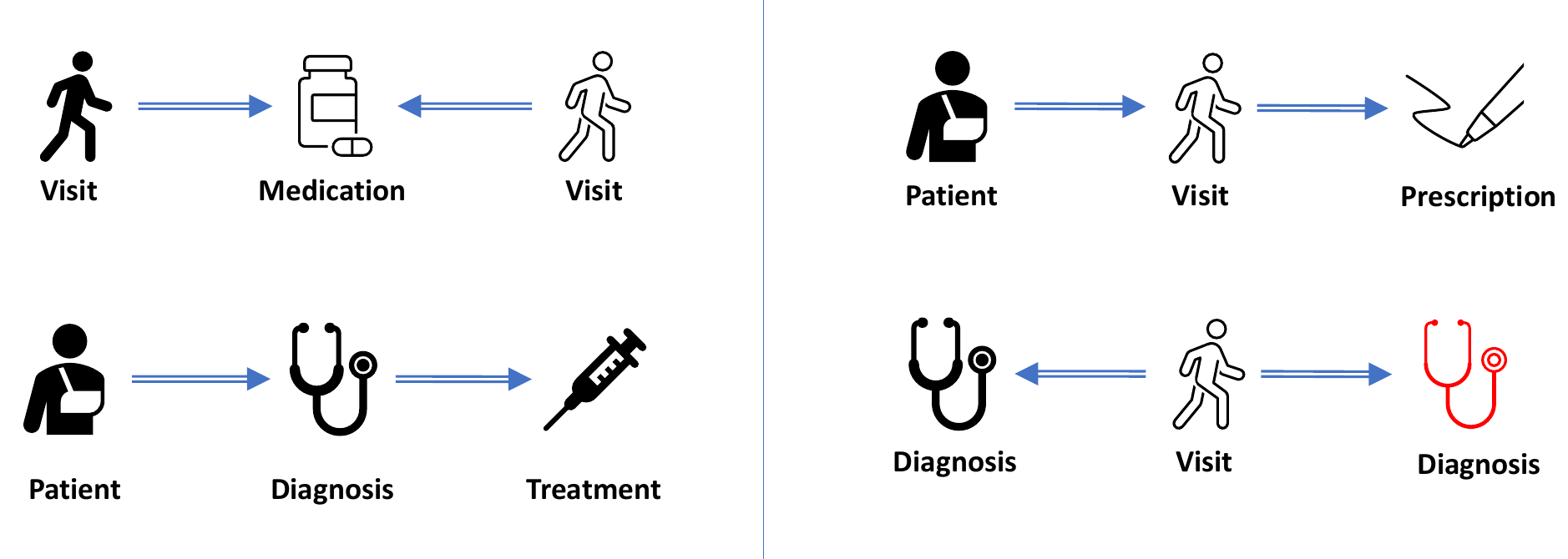}
    \caption{
    Four examples of meta-relations highlighted in the EHRs.
    These meta-relations involve structural connections between multiple nodes with different node types, highlighting the heterogeneity of the data. 
    Consequently, a homogeneous graph model is insufficient to effectively capture and represent these complex meta-relations. }
    \label{fig: meta_relations}
\end{figure}

Most of the early methods leverage the longitudinal characteristics of the clinical visits by patients and adopt recurrent neural networks (RNNs) \citep{choi2017gram, medsker2001RNN, ma2017dipole} to capture the temporal features in the EHR data. 
While these methods highlight the temporal or spatial features in the EHR data, they ignore the relational features in the EHRs --- which are potentially the most important features.
In light of the importance of relational features, there are attempts in recent research to adopt graph neural networks (GNNs) to model the EHR data.
GNNs operate on the graph domain and can highlight the relational features between the EHR entities.
Despite their successes, most of these methods operate on homogeneous graphs which focus on relations between the neighboring nodes.
This strategy does not address the heterogeneity and the semantic relations in the EHR data.
Figure \ref{fig: meta_relations} illustrates the complexity of relations between medical entities in the EHRs, highlighting the limitations of using homogeneous graphs that lead to suboptimal clinical decision performance.

Furthermore, the EHR data encompass various tasks such as drug recommendation, and predictions of in-hospital mortality, readmission, and the length of stay. 
These tasks heavily depend on the features of patients and their visits. 
However, existing designs predominantly focus on single-task prediction, and thus they fail to incorporate the multi-task characteristics of EHR data.
By sharing the learned patient or visit representations among tasks, the task-level knowledge can be leveraged to yield a better predictive performance.
Hence, a multi-task model would potentially benefit from these characteristics of EHRs.

Additionally, it is well-known that EHRs suffer from severe noise and confounding effects, as patients have diverse backgrounds and medical records (e.g., diagnosis, medications, and prescriptions).
Most of the existing methods directly operate on the features with heavy noises without adjusting for confounding effects, which hinder their performance on downstream tasks.
Hence, a denoising measure is necessary to mitigate the confounding effects in the EHRs.

Motivated by the aforementioned limitations in existing research, we propose a novel multi-task framework for EHR analysis, namely MulT-EHR (\underline{Mul}ti-\underline{T}ask EHR).
%+
MulT-EHR 
%models the EHRs with 
adpots a heterogeneous graph, which is trained by a causal denoising module and a multi-task aggregation module.
Our contributions are summarized as follows:
\begin{itemize}
    \item We propose a novel heterogeneous graph-based framework for modeling EHR data, namely MulT-EHR, which effectively mines EHR data by multi-task graph learning. 
    % proposed heterogeneous graph-based
    \item  To effectively model the structural relationships in the EHR heterogeneous graph, we first enhance the relational features within the graph by leveraging a pretraining module based on graph contrastive learning. We then adopt a transformer-based GNN architecture to effectively learn the node-level representations.
    \item We reinterpret denoising using the causal inference framework and propose a causal denoising module to adjust for the confounding effects to mitigate the catastrophically heavy noise in the EHRs. 
    % how we reinterpret existing framework
    \item We design a task-level aggregation mechanism to regularize the multi-task learning procedure by minimizing the cross-task extrapolation risk.
    This enables the single shared-weight model to leverage the cross-task knowledge more effectively.
    % learn the multi-task objectives effectively. 
    \item We perform extensive experiments on two benchmark datasets to validate the effectiveness of our method over state-of-the-art methods. 
    Our model is shown to consistently outperform the competitors over four popular clinical tasks based on EHRs --- predictions of mortality, readmission, length-of-stay, and drug recommendation.
    Enriched ablation studies demonstrate the robustness of our method to different components and hyperparameters.
\end{itemize}

%---------------------------------------------------------
\section{Related Works}
%---------------------------------------------------------

%--------------------------------
\subsection{Graph Neural Networks}
%--------------------------------
GNNs are gaining significant success in many problem domains \cite{kojima2020kgcn, hu2020HGT, liu2020hsgnn, chan2023SUMSHINE2}. 
Most of the existing GNN architectures focus on homogeneous graphs \citep{kipf2016gcn, velivckovic2017GAT, xu2018GIN, yun2019GTN}.
They learn node representations by aggregating information from the neighboring nodes on the graph topology. 
These algorithms are effective when the level of node and edge heterogeneity is negligible.
As a heterogeneous graph contains enriched semantic information, several works \citep{wang2019HAN, hu2020HGT,huang2020dahgt, yang2020MuSDAC} attempt to design GNN algorithms on heterogeneous graphs.
These works mine the complex relational information in a heterogeneous graph by designing the node-type-specific and relation-specific convolutional layers.
The research on aggregation of heterogeneous graphs is well-developed, and hence we directly use a state-of-the-art (SOTA) heterogeneous GNN architecture (i.e., the heterogeneous graph transformer \citep{hu2020HGT}) in our framework. 

%--------------------------------
\subsection{EHR Representation Learning}
%--------------------------------

Knowledge distillation from massive EHRs has been a popular topic in healthcare informatics. 
To address the longitudinal features in the EHR data, several early works \citep{ma2017dipole, ma2020concare, ma2020adacare} learned the EHR features with recurrent neural networks.
Since the EHR data represent relational information between the medical entities (e.g., patients make clinical visits), 
 %graphical models are found to be an ideal candidate learning EHR representations 
 graph-based models are widely adopted for EHR analysis
 in recent works \citep{choi2017gram, choi2018mime, ma2018kame, liu2020hsgnn}.
Early works on graph-based EHR analysis model EHR data with a homogeneous graph.
GRAM \citep{choi2017gram} is a well-known method that learns robust medical code representations by adopting a graph-based attention mechanism.
To address the heterogeneity in the EHR entities, recently there are works \citep{liu2020hsgnn, ma2018kame, zhu2021VSGNN, zhang2021grasp} attempting to model EHR data as a heterogeneous graph or a knowledge graph.
They design heterogeneous GNNs to integrate the node and edge heterogeneity in graph representation learning. 
However, almost all the existing attempts are trained on each individual
%single 
task separately, which does not incorporate the multi-task nature of the EHR data. 

%--------------------------------
\subsection{Confounding Effect Adjustment}
%--------------------------------
The challenge of addressing confounding effects in EHR data is amplified by its high degree of heterogeneity.
Variations in patients' medical histories, comorbidities, and treatment plans contribute to these confounding effects.
Consequently, there is a growing focus on learning feature representations in the counterfactual world, where confounding effects are excluded, within the healthcare research community.

Many recent works \citep{yu2019daggnn, melnychuk2022CT, hagele2022bacadi, zhao2023CausRec} take advantage of the advances in causal inference and aim to learn a representation of the data that captures the causal relationships between the variables.
By doing so, they can separate the confounding factors from the causal factors and provide a more accurate representation of the underlying causal mechanism.
CAL \citep{sui2022CAL} learns the causal features by disentangling a set of random features.
By excluding the shortcut/trivial features, it achieves promising predictive and interpretation enhancement to vanilla GNN architectures (e.g., GCN \citep{kipf2016gcn}).

%--------------------------------
\subsection{Multi-task Learning}
%--------------------------------
Multi-task learning aims to design a learning paradigm to obtain superior performance by training the tasks jointly rather than learning them independently \citep{sener2018MGDA}.
Existing works on multi-task learning can be categorized into two major trends: hard parameter sharing \citep{sener2018MGDA, dong2015MTLNLP} and soft parameter sharing \citep{yang2017traceMTL, long2017DRMMTL, xue2007DPmtl}. 
Soft parameter sharing takes all the trainable parameters task-specific but constrains them via Bayesian priors \citep{xue2007DPmtl} or a joint dictionary \citep{yang2017traceMTL, long2017DRMMTL}.
Hard parameter sharing takes a subset of parameters as shared while others remain as task-specific. 
We adopt hard parameter sharing in this paper and optimize the joint objective using an environment-invariant approach.

%--------------------------------
\section{Preliminaries}
%--------------------------------

\para{Heterogeneous Graph.} A heterogeneous graph is defined by a graph $\mathcal{G}$ = $(\mathcal{V}, \mathcal{E}, \mathcal{A}, \mathcal{R})$, where $\mathcal{V}, \mathcal{E}, \mathcal{A}$ represent the set of entities (vertices or nodes), relations (edges), and entity types respectively, 
and $\mathcal{R}$ represents the space of edge attributes. 
For $ v \in \mathcal{V}$, $v$ is mapped to an entity type by a function $\tau(v) \in \mathcal{A}$. An edge $e = (h, r, t) \in \mathcal{E}$ links the head node $h$ and the tail node $t$, and $r \in \mathcal{R}$. 
Every node $v$ has a $d$-dimensional node feature $x \in \mathcal{X}$, where $\mathcal{X}$ is the embedding space of node features.

%--------------------------------
\para{Problem: Multi-task EHR Learning.}
Given the EHR data $\mathcal{D}$, our goal is to construct a heterogeneous graph $\mathcal{G}$ from $\mathcal{D}$.
Let $\mathcal{T}_1, \ldots, \mathcal{T}_K$ on $\mathcal{G}$ be a series of $K$ tasks on $\mathcal{D}$.
We aim to train a multi-task graph neural network model $\mathcal{M}$ such that $\mathcal{M}$ can deliver high performance on $\mathcal{T}_1, \ldots, \mathcal{T}_K$.

\begin{figure}
    \centering
    \includegraphics[width=0.5\textwidth]{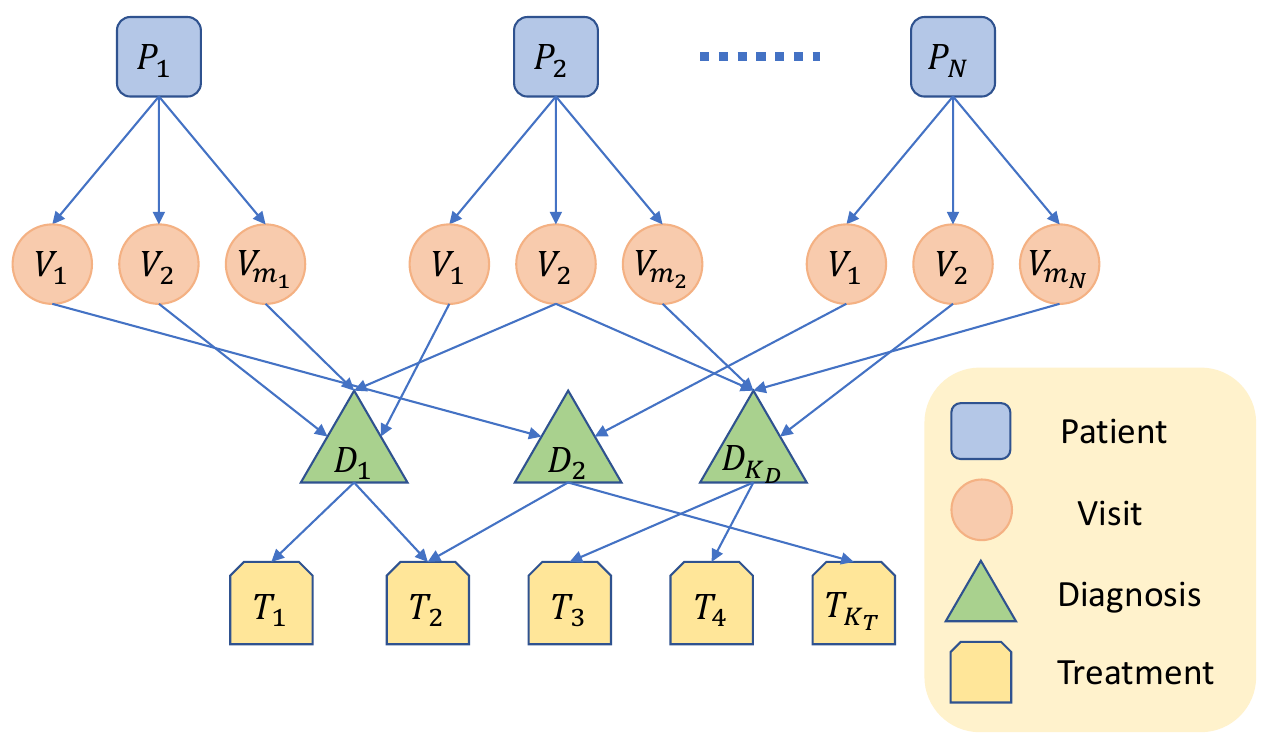}
    \caption{An example of a heterogeneous graph constructed from EHR data, where $N$ represents the total number of patients, and $m_i$ represents the total number of visits by the $i$-th patient.
    }
    \label{fig:ehr_hg_example}
\end{figure}

%---------------------------------------------------------
\section{Methodology}
%---------------------------------------------------------

\begin{figure*}[h]
    \centering
    \includegraphics[width=\textwidth]{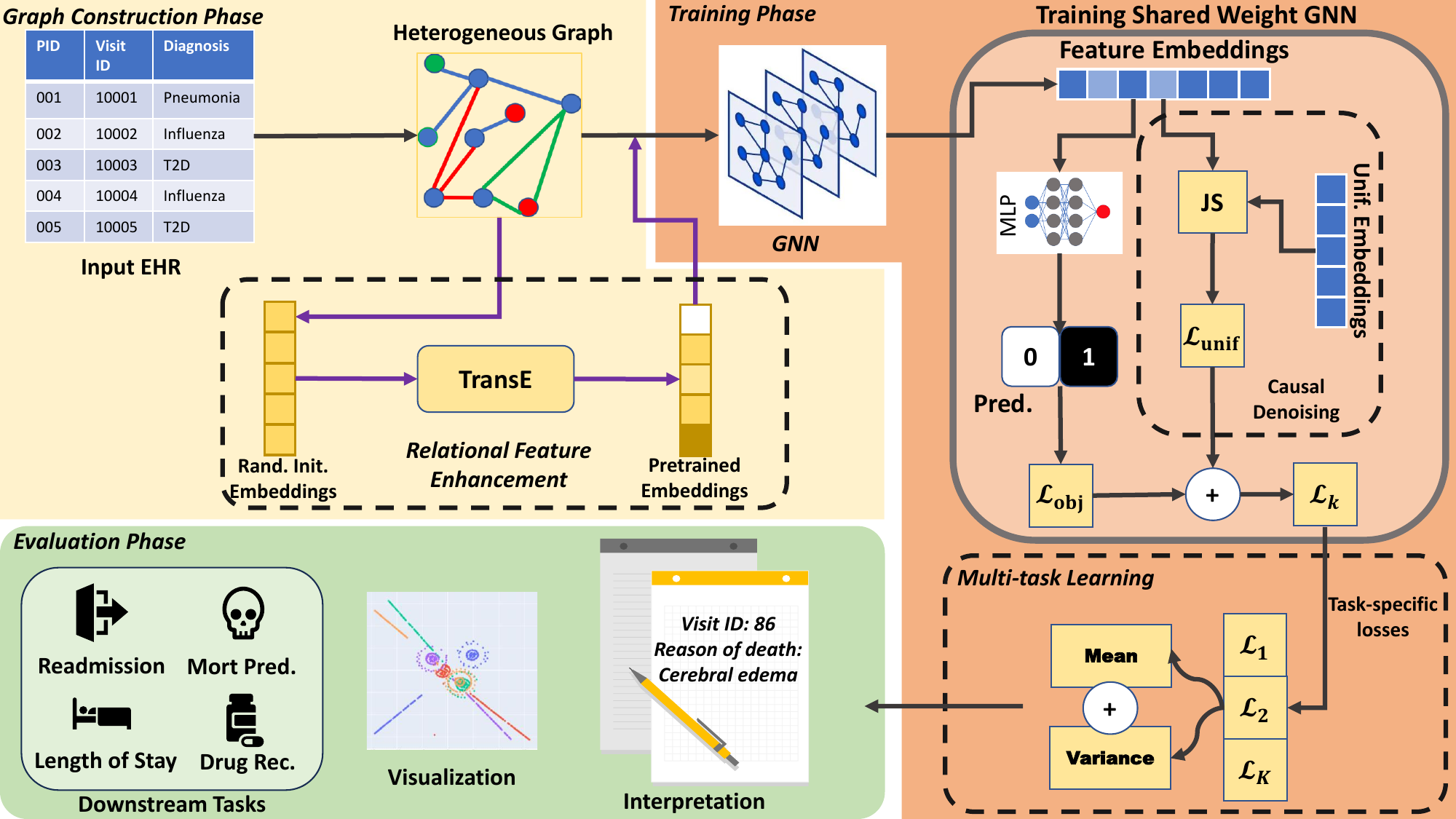}
    \caption{Overview of our proposed framework. We first construct a heterogeneous graph from the raw EHR data, and then obtain node-level representation with heterogeneous GNNs. Causal and trivial representations are disentangled and the task-specific loss is obtained by combining the classification loss $\mathcal{L}_{\rm obj}$ and the uniform loss $\mathcal{L}_{\rm unif}$.
    We adopt a task-level aggregation module to obtain the multi-task learning loss. After training the GNN, we test the GNN with different downstream tasks (e.g., mortality prediction and drug recommendation).}
    \label{fig: framework}
\end{figure*}

Our proposed framework starts with a heterogeneous graph construction stage. 
We learn the heterogeneous graph through a GNN that incorporates causal disentanglement for debiasing, which reduces the effects of confounding variables.
We then improve cross-task performance by minimizing the task-level variance. 
Figure \ref{fig: framework} illustrates the workflow of our proposed method, and Algorithm \ref{alg: silent-health} presents the training paradigm. 

%--------------------------------
\subsection{Modeling EHR with Heterogeneous Graph}
%--------------------------------
\para{Heterogeneous Graph Construction.}
We construct the heterogeneous graph by merging the tabular components in the EHR data.
We define six node types: patients, visits, diagnoses, prescriptions, procedures, and lab events. 
We further define five types of connections between the nodes: patient---visit, visit---diagnosis, visit---prescription, visit---procedure, visit---lab events. 
Figure \ref{fig:ehr_hg_example} presents the example of the heterogeneous graph constructed from EHR data.
The heterogeneous graph data structure highlights the meta-relations between the medical entities, which provides an effective data structure for mining EHRs.
Examples of the meta-relations modeled by the EHR heterogeneous graph are illustrated in Figure \ref{fig: meta_relations}.
There are visits which are indirectly connected through a common medication (upper-left panel) or one visit leads to two diagnoses (lower-right panel), and there are patients who are indirectly connected with a treatment via a specific visit (upper-right panel) or a prescription with a diagnosis (lower-left panel).
%
% By introducing the meta relations we can model the indirect relations between the EHR entities 
%
By leveraging the relational features introduced by these meta relations, we can obtain a better graph representation for EHRs and thus better performances on downstream tasks.

%--------------------------------
\para{Self-supervised Embedding Pretraining Module.}
%--------------------------------
Node features are important for optimal GNN performance.
Randomly initializing the embeddings would cause difficulties for GNN to distinguish the distributions of node embeddings, and thus might lead to trivial results.
Moreover, the randomly initialized embeddings contain no information (including the most important relational features) on the nodes, which makes learning difficult.
Hence, instead of randomly initialized node features, we pretrain the embeddings of EHRs with relational graph embedding methods, such that the relational features can be encoded into node features in this stage. 
Translational methods \citep{bordes2013transE, ji2015transD, lin2015transR} are classic approaches to translating relational features into node embeddings. 
We adopt a simple unsupervised translation method --- TransE \citep{bordes2013transE} to obtain the pretrained node embeddings,
\begin{align} \label{eq: similarity}
    f(\bm h, \bm r, \bm t) = \Vert \bm  h +  \bm  r - \bm  t \Vert,
\end{align}
where $\bm h, \bm t \in \mathbb{R}^{d}$ are the embeddings of the head and the tail of an edge, and $\bm r$ represents the embeddings corresponding to the relation type of the edge. 
We then adopt a contrastive learning-based score function to calculate the relational similarity between the nodes and backpropagate the loss to the node embeddings,
\begin{align}
    \mathcal{L}_{\rm sim} = \sum_{e \in \mathcal{G}}\sum_{e'\in S'_{e}} [f(e) - f(e') + \gamma]_+,
\end{align}
where $\gamma$ is the margin for contrastive learning, $[x]_+ = \max (x, 0)$, and $S'_{e} = \{(h', r, t) | h' \in \mathcal{V} \} \cup \{(h, r, t') | t' \in \mathcal{V} \}$ is the set of negative samples by replacing either a head $h$ or a tail $t$ with another node in the graph.
Through self-supervised learning, nodes sharing similar features would be pulled together and those whose features are different would be pushed away, leading to more distinguishable node features. 
Since most medical entities (e.g., diagnosis) are static, pretraining the node embeddings would also lead to improved inductive inference performance when new nodes (e.g., visits or patients) arrive.  

\para{Learning with Heterogeneous GNN.}
We perform node-level aggregation by adopting graph convolutional methods, which
aggregate node features by passing the information of each node to its neighboring nodes (i.e., message-passing neural networks).
However, homogeneous GNNs ignore the potential differences in node types and edge types when performing graph convolution.
To leverage the heterogeneity in the EHR graph, we adopt heterogeneous GNN architectures, where the graph convolution procedures through the layers, in general, can be formulated as
\begin{align}
    \hat{y} = \text{softmax}\left(\sum_{l=1}^L \text{act}({\rm Agg}(\mathcal{G}_l))\right),
\end{align}
where ${\rm Agg}$ is the aggregation rule,  either convolution-based (e.g., GCN \citep{kipf2016gcn}) or attention-based (e.g., GAT \citep{velivckovic2017GAT, wang2019HAN}), $\mathcal{G}_l$ is the output subgraph from layer $l$, act and softmax are the activation function and softmax normalizing module, respectively, and $\hat{y}$ is the classification probabilities output by the GNN.
%\TODO{Expand this part. Briefly describe the attention mechanism (i.e., how HGT works) and what is the final output. }
In particular, we adopt the heterogeneous graph transformer (HGT) \citep{hu2020HGT} as it yields state-of-the-art performance in predictive tasks on heterogeneous graphs.
Detailed formulation of the aggregation rule of HGT is described by \citeauthor{hu2020HGT} \citep{hu2020HGT}.
Latent representations of nodes are obtained after the aggregation. 
We then use a readout layer (e.g., multiple-layer perceptron) to obtain the prediction for each task. 
%
% Since it is difficult to fit the whole heterogeneous graph $\mathcal{G}$ into the memory, we sample a subgraph of $2000$ visits during each step. 
%
We study the effects of different GNN architectures in Section \ref{subsec: ablation GNN}.

\begin{algorithm}[t]
\begin{algorithmic}[1]
\Statex \textbf{Input:} Heterogeneous graph $\mathcal{G}$ with node features $\{H_i, \forall i \in \mathcal{V}\}$ and shared-weight GNN model $\mathcal{M}$;
\Statex Number of visits $n_{\rm visit}$ in a sampled subgraph;
\Statex Task set $ \mathcal{T} = \{\mathcal{T}_1, \ldots, \mathcal{T}_K\}$;
\Statex \textbf{Output:} The trained GNN model $\mathcal{M}$.
\State Pretraining the node features with TransE.
\For{each step}
    \State Initialize a list of losses $\mathcal{L} = \{\}$.
    \State Sample subgraph $\mathcal{G}_S$ with $n_{\rm visit}$ visit nodes;
    \For {t $\in \mathcal{T}$}
    \If {$t$ is mortality prediction}
        \State Downsample positive nodes.
    \EndIf
    \State Logits = GNN($\mathcal{G}_S$)
    \State Compute $\mathcal{L}_{\text{ce}}$ or $\mathcal{L}_{\text{bce}}$
    \State Compute uniform loss with Eq. (\ref{eq: causal-loss}) % \Comment{Pool feature with readout layer}
    \State Compute task-specific loss $\mathcal{L}_k$ with Eq. (\ref{eq: task-specific})
    \State Append $\mathcal{L}_k$ to $\mathcal{L}$
    \EndFor
    \State Compute the mean and variance of $\mathcal{L}$, and total loss with Eq. (\ref{eq: multi-task agg}).
    \State Backpropagate the loss to GNN.
\EndFor
\State \Return $\mathcal{M}$
\end{algorithmic}
\caption{The training workflow of MulT-EHR}
\label{alg: silent-health}
\end{algorithm}

%--------------------------------
\subsection{Adjusting for Confounders with Causal Inference}
%--------------------------------

\begin{figure}
    \centering
    \includegraphics[width=0.5\textwidth]{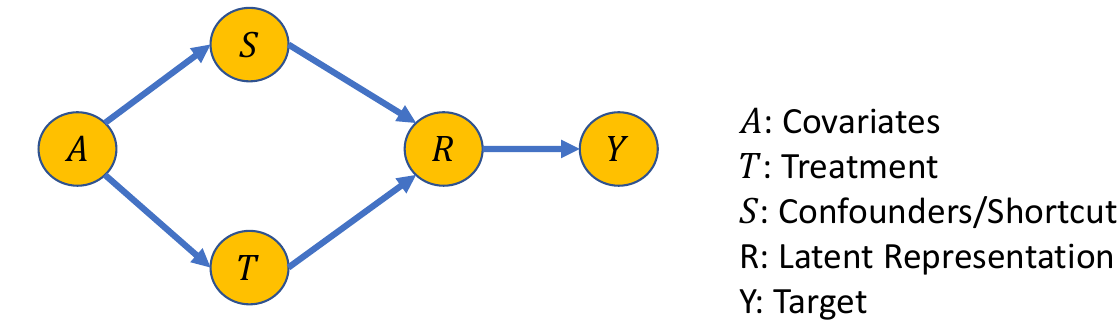}
    \caption{A causal diagram illustrating the effects of shortcut features. Without denoising, the model would make a prediction based on trivial features $S$ (i.e., the backdoor path $S$ $\to$ $R$ $\to$ $Y$)}, and the trivial variables (i.e., noise variables $S$) would hamper the prediction performance. Causal denoising aims to reduce the confounding effects by removing these backdoor paths during the training.
    \label{fig: causal_diagram}
\end{figure}

The EHR graph is known to be noisy and suffers from confounding effects. 
The trivial effects are presented as noise or shortcut features that mislead the learning process of GNNs.
Figure \ref{fig: causal_diagram} presents an illustration of the causal diagram.
The variable nodes $S$ represent the trivial features in the data, which impose noise (or confounding effects) to target prediction.
The path $A \to S \to Y$ is called the shortcut path or the backdoor path that the model would take during the forward propagation.
If there are too many shortcut paths in the graph learning process, the GNN model would be heavily affected by the shortcut (or trivial) features and this affects the learning knowledge representation in the graph.
For example, patients with presumed (but unconfirmed) interstitial lung disease may be biased toward specific or optimized imaging protocols that are intended to confirm the diagnosis, versus unsuspected cases that receive generic screening protocols \cite{ong2024shortcut}.
Here, the lung disease is the predictive variable $Y$, $R$ is the latent features which are used by the GNN to predict $Y$, and $S$ the imaging protocol is the shortcut variable.
Hence, removing the shortcuts (or backdoor paths) is critical for noise-free representation learning with GNNs.

% To minimize the effects of confounders in the graph, we disentangle the graph $\mathcal{G}$ into causal features and trivial features.
%
% \hchan{This also enables the causal interpretability of the prediction}.
%
Motivated by \citeauthor{sui2022CAL} \citep{sui2022CAL}, we introduce a causal denoising module into our framework adjusting for the confounders in the EHR data.
We first disentangle the features in $\mathcal{G}$ into two components --- the causal features and the trivial features. 
\citeauthor{sui2022CAL} \citep{sui2022CAL} proved that the causal features are invariant across training and testing distributions.
The objective of the trivial features is to match a uniform distribution to ensure the randomness of the trivial graph $\mathcal{G}_t$ \citep{sui2022CAL},
\begin{equation} \label{eq: causal-loss}
    \mathcal{L}_{\text{unif}} = \dfrac{1}{|\mathcal{D}|}\sum_{\mathcal{G} \in \mathcal{D}} \text{JS} (y_{\text{unif}}, \bm z_{\mathcal{G}_t}),
\end{equation}
where JS is the Jensen–Shannon divergence \citep{fuglede2004JSdiv} between two distributions, 
$z_{\mathcal{G}_t}$ is the trivial representation predictive with the node features from trivial graph $\mathcal{G}_t$, and $y_{\text{unif}}$ is the noise feature vector where each element is sampled from $\mathcal{U}(0, 1)$.

Learning from causal features can adjust the GNN architecture to confounding (i.e., backdoor) effects, especially when learning the EHR data with significant noise. 
\citeauthor{sui2022CAL} \citep{sui2022CAL} showed how learning through Eq. (\ref{eq: causal-loss}) can adjust for backdoor effects. 
The causal features potentially follow the counterfactual distributions which enable generalization invariability. 
Hence, the GNN model can be better generalized to the testing distributions or other tasks
%

% The binary cross-entropy loss is defined as:

% $$
% \mathcal{L}(y, \hat{y}) = -\frac{1}{N} \sum_{i=1}^N \sum_{j=1}^K y_{ij} \log(\hat{y}_{ij}) + (1 - y_{ij}) \log(1 - \hat{y}_{ij})
% $$
% where $N$ is the number of training examples, $K$ is the number of output classes, $y_{ij}$ is the true label for the $i$-th example and $j$-th output class (either 0 or 1), and $\hat{y}_{ij}$ is the predicted probability for the $i$-th example and $j$-th output class.
%
The loss functions used to train the GNN model for a single task are
\begin{align}
    \mathcal{L}_{\text{bce}} = -\frac{1}{P}\sum_{i=1}^{P} \left[ y_i \log(\sigma(z_i)) + (1 - y_i) \log(1 - \sigma(z_i)) \right],
\end{align}
\begin{align}
    \mathcal{L}_{\text{ce}} = -\frac{1}{P} \sum_{i=1}^{P} \sum_{c=1}^{C} y_{i,c} \log(\text{softmax}(z_{i,c})),
\end{align}
where \( \mathcal{L}_{\text{bce}} \) is the binary cross-entropy loss, \( \mathcal{L}_{\text{ce}} \) is the cross-entropy loss, \( y_i \) is the ground truth label for patient \( i \), \( P \) is the number of patients, \( C \) is the number of classes, and \( z_i \) and \( z_{i,c} \) are logits obtained from the model.
The final loss for task $k$ is then given by

\begin{equation} \label{eq: task-specific}
    \mathcal{L}_k = \mathcal{L}(y, \hat{y})  + \lambda \mathcal{L}_{\text{unif}},
\end{equation}
where $\mathcal{L}(y, \hat{y}) =\mathcal{L}_{\text{ce}}$ for binary classification tasks, and $\mathcal{L}(y, \hat{y}) =\mathcal{L}_{\text{bce}}$ for multi-label classification tasks, $\mathcal{L}_{\text{unif}}$ is computed by Eq. (\ref{eq: causal-loss}), and $\lambda$ is the regularization coefficient.

\begin{table*}[!t]
    \caption{Summary of MIMIC-III and MIMIC-IV datasets}
    \centering
    \scalebox{0.8}{
        \begin{tabular}{lcc|lc p{1cm}}
        \toprule
        \multicolumn{5}{c}{\textbf{MIMIC-III}} \\
        \multicolumn{1}{l}{\textbf{Node Type}} &
        \multicolumn{1}{c}{\textbf{Count}} &
        \multicolumn{1}{c}{\textbf{Avg. \# of Visits Per Entity}} &
        \multicolumn{1}{l}{\textbf{Task}} &
                \multicolumn{1}{c}{\textbf{No. Obs.}} 
        \\ \hline
        Patients  &   46,520 & --- & Mortality & 9,718 \\
        Visits &  58,976 & --- & Readmission & 9,718 \\
        Diagnoses &  6,984 & 11.04 & LoS & 44,407\\ 
        Prescriptions &  4,204 & 70.40 & Drug Recomm. & 14,142\\
        Procedures & 2,032 & 1.55 & & \\ \hline \midrule
        \multicolumn{5}{c}{\textbf{MIMIC-IV}} \\
        \multicolumn{1}{l}{\textbf{Node Type}} &
        \multicolumn{1}{c}{\textbf{Count}} &
        \multicolumn{1}{c}{\textbf{Avg. \# of Visits  Per Entity}} &
        \multicolumn{1}{l}{\textbf{Task}} &
        \multicolumn{1}{c}{\textbf{No. Obs.}} \\ 
        \hline
        Patients  &   180,733 & --- & Mortality & 125,733 \\
        Visits &  432,231 & --- & Readmission & 125,733 \\
         Diagnoses &  25,809 & 11.03 & LoS & 220,851 \\
         Prescriptions &  5,733 & 35.69 & Drug Recomm. & 147,434\\
         Procedures & 12,575 & 4.07 & & \\ \hline
        % Labevents & 710 & 377.12 & & & & 140.5224 & \\
        \bottomrule
        \end{tabular}}
    \label{tab: mimic3_summary}
\end{table*}

%--------------------------------
\subsection{Multi-task Learning via Environment-Invariant Objective}
%--------------------------------
We obtain the task-specific loss $\mathcal{L}_k$ for each task $k$ in the previous step.
We aggregate the losses from all tasks to train the single shared-weight GNN for multi-task learning.
We propose a task-invariant objective similar to \citep{wu2021EERM} to minimize the extrapolation risks in both training and testing environments.
In addition to the mean of the loss in each task, we also minimize the variance of all the $K$ losses to control the extrapolation risk,
\begin{equation} \label{eq: multi-task agg}
     \text{Var}(\{ \mathcal{L}_k  : 1 \leq k \leq K)\}) + \dfrac{\beta}{K}\sum_{k=1}^K \mathcal{L}_k,
\end{equation}
where $\text{Var}(\cdot)$ returns the variance of the set,  $\beta$ is the task-level regularization hyperparameter.
The rationale on why controlling the inter-task variance can minimize the interpolation risk is provided by \citeauthor{wu2021EERM} \citep{wu2021EERM}.

Each task can be considered as an environment that specifies a distribution of embeddings.
If the predictions for different tasks are very different, then the model may be overfitting to the current task and not learning generalizable representations.
To address this issue, the invariance objective is used in multi-task learning to encourage the model to learn task-invariant representations that are consistent across different tasks.
One way to achieve this goal is by minimizing the cross-task variance regularization term, which penalizes the model for producing very different predictions for different tasks. 
By minimizing this term, the model is encouraged to learn representations that are both task-specific and invariant across tasks, leading to better generalization performance.

%---------------------------------------------------------
\section{Experiments}
%---------------------------------------------------------

%--------------------------------
\subsection{Settings}
%--------------------------------

\para{Datasets.}
We use the MIMIC-III and MIMIC-IV datasets to evaluate our method in comparison with the competitors. 
%
% It contains a publicly available dataset of 46,520 intensive care unit (ICU) patients over 11 years.
%
Because the lab events are sparse and introduce heavy noise to the heterogeneous graph, we exclude them when constructing the graph.
Table \ref{tab: mimic3_summary} presents a summary of the types and counts of the entities in the MIMIC-III and MIMIC-IV datasets, and the details of each task.

% \begin{table*}
%     \caption{Summary of MIMIC-III and MIMIC-IV datasets}
%     \centering
%     \scalebox{0.8}{
%         \begin{tabular}{lcc|lc|lcc|lccc p{1cm}}
%         \toprule
%         \multicolumn{5}{c}{\textbf{MIMIC-III}} &
%         \multicolumn{5}{c}{\textbf{MIMIC-IV}} \\
%         \multicolumn{1}{l}{\textbf{Node Type}} &
%         \multicolumn{1}{c}{\textbf{Count}} &
%         \multicolumn{1}{c}{\textbf{Avg. \# of Visits Per Entity}} &
%         \multicolumn{1}{l}{\textbf{Task}} &
%         \multicolumn{1}{c}{\textbf{No. Obs.}} &
%         \multicolumn{1}{l}{\textbf{Node Type}} &
%         \multicolumn{1}{c}{\textbf{Count}} &
%         \multicolumn{1}{c}{\textbf{Avg. \# of Visits  Per Entity}} &
%         \multicolumn{1}{l}{\textbf{Task}} &
%         \multicolumn{1}{c}{\textbf{No. Obs.}} 
%         \\ \hline
%         Patients  &   46,520 & --- & Mortality & 9,718 & Patients  &   180,733 & --- & Mortality & 125,733\\
%         Visits &  58,976 & --- & Readmission & 9,718 & Visits &  432,231 & --- & Readmission & 125,733\\
%         Diagnoses &  6,984 & 11.04 & LoS & 44,407 & Diagnoses &  25,809 & 11.03 & LoS & 220,851\\ 
%         Prescriptions &  4,204 & 70.40 & Drug Recomm. & 14,142 &Prescriptions &  5,733 & 35.69 & Drug Recomm. & 147,434\\
%         Procedures & 2,032 & 1.55 & & & Procedures & 12,575 & 4.07\\ 
%         % Labevents & 710 & 377.12 & & & & 140.5224 & \\
%         \bottomrule
%         \end{tabular}}
%     \label{tab: mimic3_summary}
% \end{table*}

\para{Tasks and Evaluation Metrics.}
We evaluate our proposed method with common tasks on EHR data.
Our model is trained by four supervised tasks --- in-hospital mortality prediction (MORT), readmission prediction (READM), length of stay (LoS) prediction, and drug recommendation (DR).
The trained multi-task model is then evaluated on each individual task using the testing data from each task. 
%
% We report accuracy, AUROC, and macro F-1 score for the binary classification tasks and the visit-level precision@$k$, recall@$k$, and MRR@$k$, where $k = 5, 10, 20$ for the drug recommendation task.
We treat mortality prediction and readmission prediction as binary classification tasks, LoS as the multi-class classification task (with 10 classes), and drug recommendation as multi-label classification tasks (with 351 labels for MIMIC-III and 501 labels for MIMIC-IV).
We report the areas under the receiver operating curve (AUROC) and precision-recall curve (AUPR), accuracy, F1-scores, and Jaccard index for the tasks when appropriate. 
We perform five-fold cross-validation for each experiment.
Detailed definitions of the evaluation metrics are provided in the appendix.
\begin{table*}
    \caption{Performance (in \%) of our method on mortality prediction and readmission prediction on the MIMIC-III and MIMIC-IV datasets. Our proposed method, MulT-EHR, is the last row, highlighted in boldface, and standard deviations are given in brackets.}
    \centering
    \scalebox{0.85}{
        \begin{tabular}{lcc|cc|cc|cc p{1.3cm}}
        \toprule
        \multicolumn{1}{c}{}& \multicolumn{4}{c}{\textbf{Mortality Prediction}} & \multicolumn{4}{c}{\textbf{Readmission Prediction}}\\
        \multicolumn{1}{c}{}& \multicolumn{2}{c}{\textbf{MIMIC-III}} & \multicolumn{2}{c}{\textbf{MIMIC-IV}} & \multicolumn{2}{c}{\textbf{MIMIC-III}} & \multicolumn{2}{c}{\textbf{MIMIC-IV}} \\
        \multicolumn{1}{l}{\textbf{Model}} &
        \multicolumn{1}{c}{\textbf{AUROC}}& 
        \multicolumn{1}{c}{\textbf{AUPR}} &
        \multicolumn{1}{c}{\textbf{AUROC}}& 
        \multicolumn{1}{c}{\textbf{AUPR}} &
        \multicolumn{1}{c}{\textbf{AUROC}}& 
        \multicolumn{1}{c}{\textbf{AUPR}} &
        \multicolumn{1}{c}{\textbf{AUROC}}& 
        \multicolumn{1}{c}{\textbf{AUPR}} 
        \\ \hline
        GRU \citep{medsker2001RNN} & 61.09 (0.7) & 9.69 (1.5) & 67.28 (0.8) & 3.23 (0.4) & 65.58 (1.1) & 68.57 (1.6) & 68.46 (0.5) & 68.85 (0.6) \\
        Transformer \citep{hochreiter1997LSTM} & 62.23 (3.0) & 10.70 (1.0) & 64.79 (1.2) & 2.85 (0.2) & 63.70 (0.5) & 68.92 (1.2) & 67.74 (0.9) & 69.23 (1.1)\\
        Deepr \citep{nguyen2016deepr} & 58.61 (2.4) & 11.87 (0.4) & 65.13 (0.8) & 3.20 (0.3) & 65.10 (1.3)  & 68.68 (1.2) & 67.20 (0.5) & 68.06 (0.3)\\
        GRAM \citep{choi2017gram} & 60.00 (1.0) & 11.00 (1.0) & 65.00 (1.0) & 4.00 (0.0) & 64.00 (0.0) & 67.00 (1.0) & 66.00 (0.0) & 66.00 (0.0)\\
        Concare \citep{ma2020concare} & 61.98 (1.8) & 9.67 (1.5) & 65.37 (1.9) & 3.15 (0.2) & 65.28 (1.1) & 66.67 (1.9) & 68.67 (0.1) & 69.60 (0.7) \\
        Dr. Agent \citep{gao2020dragent} & 57.52 (0.4) & 9.66 (0.8)  & 64.59 (1.4) & 3.47 (0.3) & 64.86 (2.6) & 67.41 (1.0) & 68.09 (0.3) & 69.24 (0.6)\\ 
        AdaCare \citep{ma2020adacare} & 64.84 (2.3) & 12.21 (1.6) & 65.64 (0.1) & 3.30 (0.1) & 64.90 (0.6) & 67.49 (0.7) & 67.64 (0.3) & 67.91 (0.3) \\ 
        StageNet \citep{gao2020stagenet} &  64.49 (0.7) & 16.67 (3.0) & 65.58 (1.9) & 3.15 
        (0.2) & 62.38 (0.4) & 68.05 (0.8) & 67.81 (0.6) & 68.29 (0.5) \\
        GRASP \citep{zhang2021grasp} & 59.29 (3.2) & 9.32 (1.9) & 64.48 (1.9)  & 2.69 (0.2) & 66.91 (1.6) & 70.41 (1.6) & 67.34 (0.4) & 67.17 (0.8)\\ 
        GraphCare \citep{jiang2023graphcare} & 70.00 (1.0) & 16.00 (0.0) & 70.50 (0.7) & 4.90 (0.1) & 68.90 (0.4) & 70.80 (0.8) & 68.00 (0.2) & 67.10 (0.4)\\
        %\multicolumn{1}{c}
        {\textbf{MulT-EHR}} & \textbf{71.28 (0.6)} & \textbf{17.48 (2.5)} & \textbf{70.80 (1.8)} & \textbf{5.15 (0.2)} & \textbf{71.33 (0.4)} & \textbf{71.23 (0.1)} & \textbf{68.77 (0.2)} & \textbf{70.26 (0.2)}\\ \hline \bottomrule
        \end{tabular}}
    \label{tab: results_MIMIC3}
\end{table*}
%--------------------------------
\subsection{Implementation Details}
%--------------------------------

The proposed framework is implemented in Python with the \textit{Pytorch} library on a server equipped with four NVIDIA TESLA V100 GPUs. 
We use the \textit{dgl} library to perform graph-related operations, and \textit{pyhealth} \citep{zhao2021pyhealth} to benchmark SOTA methods and perform EHR-related operations.
The dropout ratio of each dropout layer is set as 0.2. 
All models are trained with 1000 epochs with early stopping. 
We choose the model at the epoch where it yields the best performance in terms of AUROC.
We adopt the cross-entropy loss to train the network for classification tasks, and MSE for regression tasks.
We use the Adam optimizer to optimize the model with a learning rate of $5 \times 10^{-5}$ and a weight decay of $1 \times 10^{-5}$. 
We perform data augmentations on the training graphs by randomly dropping the edges and nodes, and adding Gaussian noises to the node and edge features.

\para{Temperature Annealing. } We are aware of the vanishing classification loss in practice. Therefore, we alleviate this issue by annealing the temperature over the training epochs with the schedule $\tau = \max (0.05, \exp(rp))$, where $p$ is the training epoch and $r = 0.01$. 

\para{Subgraph Sampling. } Since it is not always possible to pass the whole EHR graph into the memory (especially for MIMIC-IV), we compose subgraphs by sampling $n_{\rm visit}$ visits nodes and their connected nodes at each epoch.
We set $n_{\rm visit}=2000$ as this parameter is fine-grained with empirical experience to which the performance is less sensitive. 

\para{Downsampling for MORT Task. }
We are aware that the samples in the mortality prediction task are heavily imbalanced (i.e., most of the samples are alive).
We therefore perform downsampling during the training to balance the samples.

\begin{table*}[]
    \caption{Performance of our method on prediction of the length of stay on the MIMIC-III and MIMIC-IV datasets. Our proposed method, MulT-EHR, is the last row, highlighted in boldface, and standard deviations are shown in brackets.}
    \centering
    \resizebox{0.85\linewidth}{!}{%
        \begin{tabular}{lccc|ccc p{1.3cm}}
        \toprule
        \multicolumn{1}{c}{}& \multicolumn{6}{c}{\textbf{Prediction of Length of Stay}} \\
        \multicolumn{1}{c}{}&
        \multicolumn{3}{c}{\textbf{MIMIC-III}}  &
        \multicolumn{3}{c}{\textbf{MIMIC-IV}}\\
        \multicolumn{1}{l}{\textbf{Model}} &
        \multicolumn{1}{c}{\textbf{Accuracy}} &
        \multicolumn{1}{c}{\textbf{AUROC}}& 
        \multicolumn{1}{c}{\textbf{F1}} &
        \multicolumn{1}{c}{\textbf{Accuracy}} &
        \multicolumn{1}{c}{\textbf{AUROC}}& 
        \multicolumn{1}{c}{\textbf{F1}} \\ \hline
        GRU \citep{medsker2001RNN} &  42.14 (0.6) & 80.23 (0.2)  &  27.36 (0.7)  & 38.30 (0.4) & 81.23 (0.1) & 32.03 (0.3)\\
        Transformer \citep{vaswani2017transformer} & 41.68 (0.7) & 79.30 (0.8) & 27.52 (0.8) & 37.40 (0.3) & 80.73 (0.5) & 31.86 (0.5) \\
        Deepr \citep{nguyen2016deepr} & 39.31 (1.2) & 78.02 (0.4) & 25.09 (1.3) & 36.00 (0.9) & 80.53 (0.3) & 31.05 (0.7)  \\
        GRAM \citep{choi2017gram} & 40.00  (0.0) & 78.00  (0.0) & 34.00 (0.0) & 35.00 (0.0) & 79.00 (0.0) & 32.00 (0.0)\\
        Concare \citep{ma2020concare} & 42.04 (0.6) & 80.27 (0.3) & 25.44 (1.3) & 37.80 (0.4) & 81.04 (0.3) & 30.41 (0.4)\\
        Dr. Agent \citep{gao2020dragent} & 41.40 (0.5) & 79.45 (0.6) & 27.55 (0.3) &  38.15 (0.1) & 81.01 (0.1) & 31.25 (0.5) \\ 
        AdaCare \citep{ma2020adacare} & 40.7 (0.8) & 78.73 (0.4) & 26.26 (0.8) & 37.27 (0.6) & 80.63 (0.2) & 31.19 (0.2) \\ 
        StageNet \citep{gao2020stagenet} & 40.18 (0.7) &  77.94 (0.2)  & 26.63 (1.2) & 36.47 (0.7) &  80.04 (0.2) & 30.55 (0.8)  \\
        GRASP \citep{zhang2021grasp} & 40.66 (0.3) &  78.97 (0.4) & 22.80 (0.8) & 35.28 (0.1) & 79.86 (0.4) & 26.95 (2.8) \\
        GraphCare \citep{jiang2023graphcare} & 42.00 (0.0) & 80.00 (0.0) & 37.00 (0.0) & 36.00 (0.0) & 80.00 (0.0) & 32.00 (0.0)  \\
        % SafeDrug \citep{yang2021safedrug} &  &  &  \\
        %\multicolumn{1}{c}
        {\textbf{MulT-EHR}} & \textbf{43.17 (1.1)} & \textbf{81.53 (0.1)} & \textbf{41.44 (0.6)}  &  \textbf{41.40 (0.4)} &  \textbf{84.04 (0.2)} & \textbf{38.46 (0.4)}\\ \hline
        \bottomrule
        \end{tabular}
    }    
    \label{tab: results_MIMIC3_LOS}
\end{table*}

% \begin{table}[!ht]
% \caption{Datasets Summary}
%     \centering
%     \resizebox{0.95\linewidth}{!}{\def\arraystretch{1}
% \begin{threeparttable}
% \begin{tabular}{lcccc}
% \toprule
% \multicolumn{1}{l}{\textbf{Dataset}} &
% \multicolumn{1}{c}{No. Classes} &
% \multicolumn{1}{c}{No. Nodes} &
% \multicolumn{1}{c}{No. Training} &
% \multicolumn{1}{c}{No. Testing}\\ \hline
% \toprule
% \multicolumn{1}{l}{\textbf{Twich-explicit\tnote{1}}} & 2 & & 3,629 & 1,725\\ \hline
% \multicolumn{1}{l}{\textbf{DBLP\tnote{2}}} &  1 & 1904 & 784\\ \hline
% \bottomrule
% \end{tabular}
% % \begin{tablenotes}[flushleft]\footnotesize
% % \end{tablenotes}
% \end{threeparttable}}
% \label{tab:Datasets summary}
% \end{table}

%--------------------------------
\subsection{Comparable Methods}
%--------------------------------
We compare our method to the following competitors: GRU \citep{medsker2001RNN}, Transformer \citep{vaswani2017transformer}, GRAM \citep{choi2017gram}, StageNet \citep{gao2020stagenet}, AdaCare \citep{ma2020adacare}, Concare \citep{ma2020concare}, GRASP \citep{zhang2021grasp}, Deepr \citep{nguyen2016deepr}, and GraphCare \citep{jiang2023graphcare}.
For the drug recommendation task, we further include the following competitors which are distinctively designed to tackle this task: MICRON \citep{yang2021micron}, Safedrug \citep{yang2021safedrug} and MoleRec \citep{yang2023molerec}.
Detailed description of each baseline method can be found in the appendix. 
\begin{table*}[h]
    \caption{Performance of our method on drug recommendation on the MIMIC-III and MIMIC-IV datasets. Our proposed method, MulT-EHR, is the last row, highlighted in boldface, and standard deviations are given in brackets.}
    \centering
     \resizebox{0.85\linewidth}{!}{%
        \begin{tabular}{lccc|ccc p{1.3cm}}
        \toprule
        \multicolumn{1}{c}{}& \multicolumn{3}{c}{\textbf{MIMIC-III}} & \multicolumn{3}{c}{\textbf{MIMIC-IV}}\\
        \multicolumn{1}{l}{\textbf{Model}} &
        \multicolumn{1}{c}{\textbf{AUROC}} &
        \multicolumn{1}{c}{\textbf{AUPR}}& 
        \multicolumn{1}{c}{\textbf{Jaccard}} &
        \multicolumn{1}{c}{\textbf{AUROC}} &
        \multicolumn{1}{c}{\textbf{AUPR}}& 
        \multicolumn{1}{c}{\textbf{Jaccard}} \\ \hline
        GRU \citep{medsker2001RNN} & 96.38 (0.1) & 64.75 (0.2) & 45.78 (0.5) & 97.67 (0.1) & 63.99 (1.3) & 45.32 (0.3) \\
        Transformer \citep{vaswani2017transformer} & 95.87 (0.0) & 60.19 (0.1)  & 41.14 (0.4) & 97.65 (0.1) & 63.79 (1.2) & 44.88 (0.9) \\
        GRAM \citep{choi2017gram} & 94.00 (0.0) & 77.00 (0.0) & 48.00 (0.0) & 94.00 (0.0) & 60.00 (0.0) & 45.00 (0.0)\\
        Concare \citep{ma2020concare} & 95.78 (0.1) & 61.67 (0.3) & 43.43 (0.5) & 97.43 (0.0) & 61.60 (0.1) & 43.51 (0.8)  \\
        Deepr \citep{nguyen2016deepr}  & 96.09 (0.2) & 62.48 (0.0) & 44.45 (0.4) & 97.35 (0.0) & 60.31 (0.1) & 43.31 (0.7)\\
        Dr. Agent \citep{gao2020dragent} & 96.41 (0.1) & 64.16 (0.5) & 44.09 (0.7) & 97.67 (0.8) & 64.13 (0.1) & 44.52 (0.6) \\ 
        AdaCare \citep{ma2020adacare} & 95.86 (0.0) & 60.76 (1.1)  &  42.00 (1.3) & 97.53 (0.0)  & 62.54 (0.0) & 43.51 (0.4) \\ 
        StageNet \citep{gao2020stagenet} & 96.05 (0.3) & 62.43 (2.4) & 44.60 (3.7) & 97.62 (0.1) & 63.94 (1.5) & \textbf{45.69 (2.1)}\\
        GRASP \citep{zhang2021grasp} & 96.01 (0.1) & 62.53 (0.8)  & 44.12 (0.5) & 97.44 (0.0) & 61.73 (0.3) & 42.47 (0.7) \\
        GraphCare \citep{jiang2023graphcare} & 95.00 (0.0) & 77.80 (0.2) & 48.20 (0.4) & 92.00 (0.0) & 69.00 (0.0) & 41.00 (0.0) \\ \hline
        MICRON \citep{yang2021micron} & 96.21 (0.0) & 63.84 (0.1) & 45.95 (0.5) & 97.57 (0.0) & 63.19 (0.1) & 45.35 (0.1)\\
        MoleRec \citep{yang2023molerec} & 92.00 (0.1) & 69.80 (0.3) &  43.10 (0.3) &92.10 (0.1) & 68.60 (0.1) & 40.60 (0.3) \\
        SafeDrug \citep{yang2021safedrug} & 94.20 (0.1) & 76.40 (0.0) & 47.20 (0.4) &91.80 (0.2)  & 66.40 (0.5)  &  44.30 (0.3) \\ \hline
        %\multicolumn{1}{c}
        {\textbf{MulT-EHR}} & \textbf{96.67 (0.1) } & \textbf{78.58 (0.2)} & \textbf{52.20 (0.8)} & \textbf{97.68 (0.1)} & \textbf{70.43 (0.2)} & 44.23 (0.0)\\ \hline
        \bottomrule
        \end{tabular}
    }
    \label{tab: results_MIMIC3_drug}
\end{table*}
%--------------------------------
\subsection{Quantitative Results}
%--------------------------------

Tables \ref{tab: results_MIMIC3}--\ref{tab: results_MIMIC3_drug} present the results of different tasks on the MIMIC-III and MIMIC-IV datasets. 
We observe that our proposed framework outperforms the competitive methods on all tasks, which validates its predictive performance. 
Our method adopts one model for all benchmark tasks, which does not require training a GNN for each individual task. 
Remarkably, despite using a single-shared weight model, our approach consistently outperforms single-task methods across all individual tasks.
We also observe that for the drug recommendation task, our method not only outperforms the SOTA methods for EHR prediction but also recent methods \citep{yang2021micron, yang2023molerec, yang2021safedrug} specifically tackling the drug recommendation tasks.
This compelling evidence suggests that through multi-task learning, we exert the potential to surpass the limitations of single-task models by leveraging knowledge from other downstream tasks.
% For drug recommendation, our method can outperform the methods specifically designed for this task, such as Safedrug \citep{yang2021safedrug}.
%
Our model can even consistently outperform the large language model-based methods (e.g., GraphCare \citep{jiang2023graphcare}) in the downstream tasks, where these methods borrow excessive knowledge from the open-world knowledge base.

%
% \hchan{This validates that our proposed multi-task learning module can potentially improve performance on individual tasks by leveraging cross-task knowledge.}

%--------------------------------
\subsection{Qualitative Evaluation}
%--------------------------------

\para{Embedding Visualization.}
We visualize the node embeddings of each type of entity to evaluate the performance of feature representation learning.
\begin{figure*}[h]
    \centering
    \includegraphics[width=0.32\textwidth]{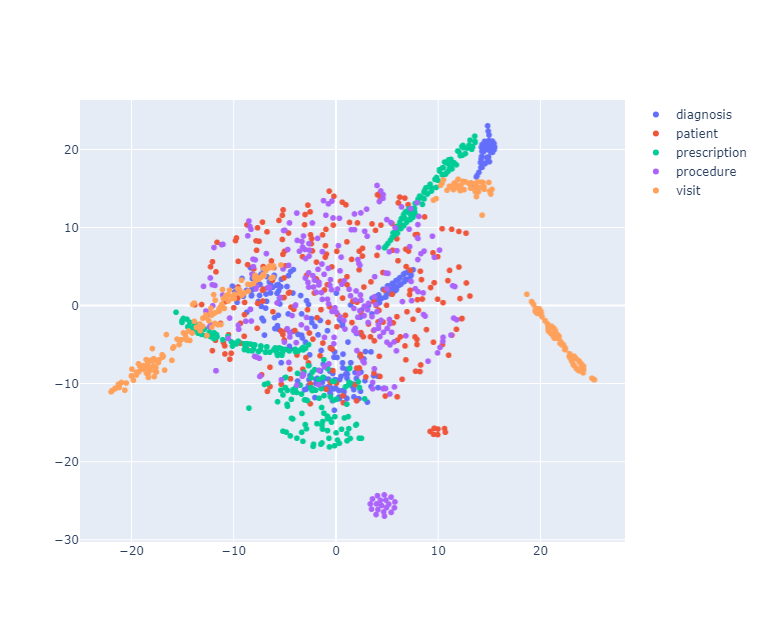}
    \includegraphics[width=0.32\textwidth]{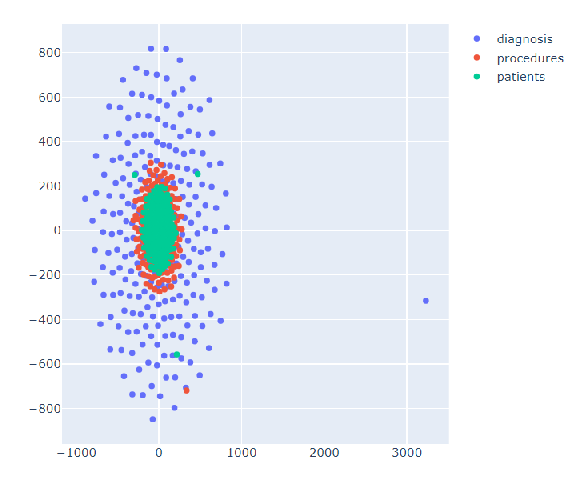}
    \includegraphics[width=0.32\textwidth]{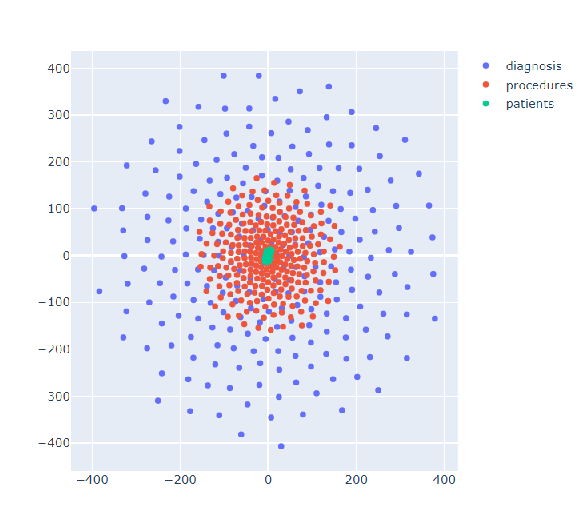}
    \caption{T-SNE scatter plot of the embeddings (Left panel: MulT-EHR, Middle panel: StageNet \citep{gao2020stagenet}, Right panel: Deepr \citep{nguyen2016deepr}). We observe that MulT-EHR can capture more complex patterns than other methods.
    }
    \label{fig: embeddings}
\end{figure*}
Figure \ref{fig: embeddings} presents the T-SNE (t-distributed stochastic neighbor embedding) plots of the embeddings generated by different methods. 
In general, the embeddings are clustered according to their node types, which validates that the embeddings can learn the unique representation of each node type. 
We also compare the embeddings across different methods. 
We observe that all the methods can capture the patterns of medical entities. 
However, the pattern captured by MulT-EHR is more unique and complex than other methods.
This shows that our method is more capable of capturing the unique pattern presented in the EHR data.

We also compare the embeddings from different GNN architectures.
We observe that the embeddings generated by heterogeneous GNNs have more unique patterns than those generated by homogeneous GNNs.
This also validates that modeling EHRs by heterogeneous graphs can potentially learn the more complex relationships and patterns in the EHR data.
Visualizations of embeddings generated by other GNN architectures can be found in the appendix.

%--------------------------------
\para{Prediction Interpretations.}
%--------------------------------
We perform a case study on a specific visit of a patient to evaluate the decision process on readmission prediction.
For the selected visit node, we select the top 3 diagnosis edges that have the highest edge attention scores.
Table \ref{tab: case_study} presents the visit and the selected diagnoses associated with their readmission prediction (together with the ICD9 codes and description of the diagnoses).
We observe that our model can select diagnoses related to brain functionality and cancer according to their attention scores, which provides evidence that the model can effectively capture the semantic information in EHRs when making readmission predictions.

% \begin{table*}
%     \centering
%     \caption{Case study on the diagnoses that are highly associated with the readmission of an arbitrarily selected visit. We observe that the selected diagnoses of high attention scores are all related to brain functionality and cancer, which provides evidence that the model can effectively capture the semantic information in the EHR.}
%     \begin{tabular}{c|clcc}
%     \toprule
%          &  \multicolumn{3}{c}{\textbf{Predicted Diagnosis}} & \\ 
%          \textbf{Case}& ICD9 Code & \multicolumn{1}{l}{Description} &  Attention Score  \\ \hline
%          \multirow{3}{*}{Visit 86} & 1983 & Secondary malignant neoplasm of brain and spinal cord  &  7.4292 \\
%          & 3485 & Cerebral edema & 7.3845 \\
%          & V5865 & Long-term (current) use of steroids & 7.3499\\
%          \bottomrule
%     \end{tabular}
%     \label{tab: case_study}
% \end{table*}

\begin{table}
    \centering
    \caption{Case study on the diagnoses that are highly associated with the readmission of an arbitrarily selected visit. We observe that the selected diagnoses of high attention scores are all related to brain functionality and cancer, which provides evidence that the model can effectively capture the semantic information in the EHR data.}
    \begin{tabular}{c|clc}
    \toprule
         &  \multicolumn{3}{c}{\textbf{Diagnoses}}  \\ 
         \textbf{Case}& ICD9 & \multicolumn{1}{l}{Description} &  Att. Score  \\ \hline
         \multirow{7}{*}{Visit 86} & \multirow{3}{*}{1983} & Secondary malignant &  \multirow{3}{*}{7.4292} \\ 
         & &  neoplasm of brain \\
         & &  and spinal cord  \\ \cline{2-4}
         & \multirow{2}{*}{3485} & Cerebral  & \multirow{2}{*}{7.3845} \\
         & & Edema \\ \cline{2-4}
         & \multirow{2}{*}{V5865} & Long-term (current)  & \multirow{2}{*}{7.3499}\\
         & & use of steroids\\\hline
         \bottomrule
    \end{tabular}
    \label{tab: case_study}
\end{table}

% The causal disentanglement highlights causal features in the heterogeneous graph which enables interpretations on nodes.
%
% We select the nodes that have the largest causal features (i.e., largest signal) on the multi-task GNN to validate the interpretability of our method.
% We leverage the information-based explanation methods \citep{schwab2019cxplain} to compute the contribution of each node. 
%
% The contribution of each node is defined by the information loss (in the difference in classification entropy) when excluding them from the graph.
%
% This contribution is also known as Granger causality \citep{granger1969GrangerCausality1}.

% Given a trained GNN model $\mathcal{M}$, the causal contribution of each node $v$ is given by 
% \begin{equation} \label{eq:causal explain}
%     \Delta_{\delta, v} = \mathcal{L}(y, \Tilde{y}_{\mathcal{G}}) - \mathcal{L}(y, \Tilde{y}_{\mathcal{G} \text{\textbackslash} \{ v\}}),
% \end{equation}
% where $y$ is the true label and $\Tilde{y}_{\mathcal{G}} = \mathcal{M}(\mathcal{G})$ and $\Tilde{y}_{\mathcal{G} \text{\textbackslash} \{ v\}} = \mathcal{M}(\mathcal{G} \text{\textbackslash} \{ v\})$ are the predicted labels from $\mathcal{M}$ with input graphs $\mathcal{G}$ and $\mathcal{G} \text{\textbackslash}\{ v\}$, respectively. 
% \TODO{Evaluate the explanation with and without causal feature, plus the explanation on noise features}

%---------------------------------------------------------
\section{Ablation Analysis}
%---------------------------------------------------------
%--------------------------------
\subsection{Ablation Study on Different Components}
%--------------------------------

To validate the contributions of each component of our model, we deactivate the causal debiasing and/or multi-task learning modules to examine their effects on the results.
Table \ref{tab: components} presents the results on the readmission task. 
We observe that including either the causal denoising module or the multi-task aggregation module leads to improvement in performance, while including both modules results in the best performance.
This validates both modules proposed in our framework improve the learning performance. 

\begin{table}
    \caption{Effects of the causal denoising module and task-level aggregation module.
    We evaluate the performance on the MIMIC-III readmission task.
    }
    \centering
        \begin{tabular}{cc|cccc p{1.2cm}}
        \toprule
        \multicolumn{1}{l}{\textbf{Causal}} &
        \multicolumn{1}{c}{\textbf{Task-Agg.}} &
        \multicolumn{1}{c}{\textbf{AUROC}} &
        \multicolumn{1}{c}{\textbf{AUPR}} &
        \multicolumn{1}{c}{\textbf{F1}}
        \\ \hline
          &   & 61.67 & 63.56 & 67.55 \\
        \checkmark &  & 63.91 & 68.85 & 68.89\\
         &  \checkmark & 67.65 & 68.93 &69.23\\
         \checkmark &  \checkmark & \textbf{71.33} & \textbf{70.61} & \textbf{69.86}\\\hline
        \bottomrule
        \end{tabular}
    \label{tab: components}
\end{table}

%--------------------------------
\subsection{Effects of Different Numbers of Tasks}
%--------------------------------
We show that as more tasks are incorporated into our multi-task learning method, the predictive task performance can be improved due to cross-task knowledge sharing. 
We experiment with one to four tasks, and Table \ref{tab: tasks} presents the results.
We observe that as the number of tasks in the training increases, the performance on readmission prediction improves accordingly. 
This validates that our multi-task learning framework can leverage more inter-task knowledge as more tasks are included in the training stage.

\begin{table}
    \caption{
    Performance of our framework on the different numbers of prediction tasks (R: readmission; M: mortality; D: drug recommendation; L: length of stay).
    We benchmark the performance on the readmission prediction task.
    }
    \centering
    \scalebox{0.8}{
        \begin{tabular}{lccc|ccc p{1.2cm}}
        \toprule
        \multicolumn{1}{l}{\textbf{}} &
        \multicolumn{3}{c}{\textbf{MIMIC-III}} & 
        \multicolumn{3}{c}{\textbf{MIMIC-IV}}
        \\
        \multicolumn{1}{l}{\textbf{Tasks}} &
        \multicolumn{1}{c}{\textbf{AUROC}} &
        \multicolumn{1}{c}{\textbf{AUPR}} &
        \multicolumn{1}{c}{\textbf{F1}} &
        \multicolumn{1}{c}{\textbf{AUROC}} &
        \multicolumn{1}{c}{\textbf{AUPR}} &
        \multicolumn{1}{c}{\textbf{F1}}
        \\ \hline
         R &  62.89 & 69.37 & 66.38 &  68.30 & 69.44 & 68.32\\
         RM & 65.39 & 67.29 &  68.02 & 68.08 & 68.94 & 68.40\\
         RD & 62.89 & 65.80 & 68.03& 68.71 & 69.91 & 67.68\\
         RMD &  68.87 & 69.57 & 69.76 & 66.50 & 66.77 & 67.86 \\
         RDL &  65.59 & 66.79 & 69.27 & 67.63 & 69.65& 65.64\\
         RMDL & \textbf{71.33} & \textbf{70.61} & \textbf{69.86} & \textbf{68.74} & \textbf{70.02 }& \textbf{68.45}\\ \hline
        \bottomrule
        \end{tabular}}
    \label{tab: tasks}
\end{table}

\begin{table*}
    \caption{Effects of different GNN architectures by evaluating the performance of the readmission prediction task on the MIMIC-III dataset.
}
    \centering
        \begin{tabular}{lccc|ccc p{1.2cm}}
        \toprule
                \multicolumn{1}{l}{\textbf{}} &
        \multicolumn{3}{c}{\textbf{MIMIC-III}} & 
        \multicolumn{3}{c}{\textbf{MIMIC-IV}}
        \\
        \multicolumn{1}{l}{\textbf{Model}} &
        \multicolumn{1}{c}{\textbf{AUROC}} & 
        \multicolumn{1}{c}{\textbf{AUPR}} & 
        \multicolumn{1}{c}{\textbf{F1}} &
        \multicolumn{1}{c}{\textbf{AUROC}} & 
        \multicolumn{1}{c}{\textbf{AUPR}} & 
        \multicolumn{1}{c}{\textbf{F1}} 
        \\ \hline
        GCN \citep{kipf2016gcn}& 56.26 & 58.55 & 67.42 & 57.62 & 57.90 & 61.05 \\
        GAT \citep{velivckovic2017GAT} & 68.34 & 69.86 & 66.43 & 58.58 &  57.31 & 66.83 \\
        GIN \citep{xu2018GIN} & 60.47 & 64.23 & 67.59 & 64.54 & 65.14 & 66.95 \\
        % HAN \citep{wang2019HAN} &  &   & \\ 
        RGCN \citep{schlichtkrull2018RGCN} & 52.75& 55.53 & 64.66 & 51.25 & 51.64 & 63.60\\ 
        HGT \citep{hu2020HGT} & \textbf{71.33} & \textbf{70.61} & \textbf{69.86} & \textbf{68.74} & \textbf{70.02} & \textbf{68.31} \\ \hline
        \bottomrule
        \end{tabular}
    \label{tab: ablations_aggregation}
\end{table*}

\begin{table}[h]
    \caption{
    Performance of our framework on different numbers of layers $L$.
    We evaluate the performance of the readmission prediction task on MIMIC-III and MIMIC-IV.
    }
    \centering
    \scalebox{0.85}{
        \begin{tabular}{lccc|ccc p{1.2cm}}
        \toprule
        \multicolumn{1}{l}{} & \multicolumn{3}{c}{\textbf{MIMIC-III}} & \multicolumn{3}{c}{\textbf{MIMIC-IV}}\\
        \multicolumn{1}{l}{$L$} &
        \multicolumn{1}{c}{\textbf{AUROC}} &
        \multicolumn{1}{c}{\textbf{AUPR}} &
        \multicolumn{1}{c}{\textbf{F1}}&
        \multicolumn{1}{c}{\textbf{AUROC}} &
        \multicolumn{1}{c}{\textbf{AUPR}} &
        \multicolumn{1}{c}{\textbf{F1}}
        \\ \hline
        1 & 64.08 & 68.80  & 66.81 & 67.24 &68.02 & 65.87\\
        2 & \textbf{71.33} & \textbf{70.61} & \textbf{69.86} &  \textbf{68.74} & \textbf{70.02} & \textbf{68.31}\\
        3 & 67.61 & 70.44 &68.56 & 68.08 & 69.74 & 68.06\\
        4 & 66.42 & 70.34 & 67.32 & 68.33 & 69.11 & 68.08 \\ \hline
        \bottomrule
        \end{tabular}}
    \label{tab: no-layers}
\end{table}

%--------------------------------
\subsection{Effects of Different GNN Architetures} \label{subsec: ablation GNN}
%--------------------------------

We compare different graph convolutional methods to show how ablations in aggregation methods affect our framework. 
Table \ref{tab: ablations_aggregation} presents the results of the MIMIC-III hospital readmission task. 
We observe that our method is overall robust when the GNN architecture changes. 
However, using a homogeneous GNN architecture (e.g., GCN) would hamper the predictive performance since they only consider direct connections in the graph by neighbour averaging.
Hence, they cannot leverage the structural information in the EHR data, which leads to less satisfactory performance.

\begin{figure}[h]
\centering
\includegraphics[width=0.46\textwidth]{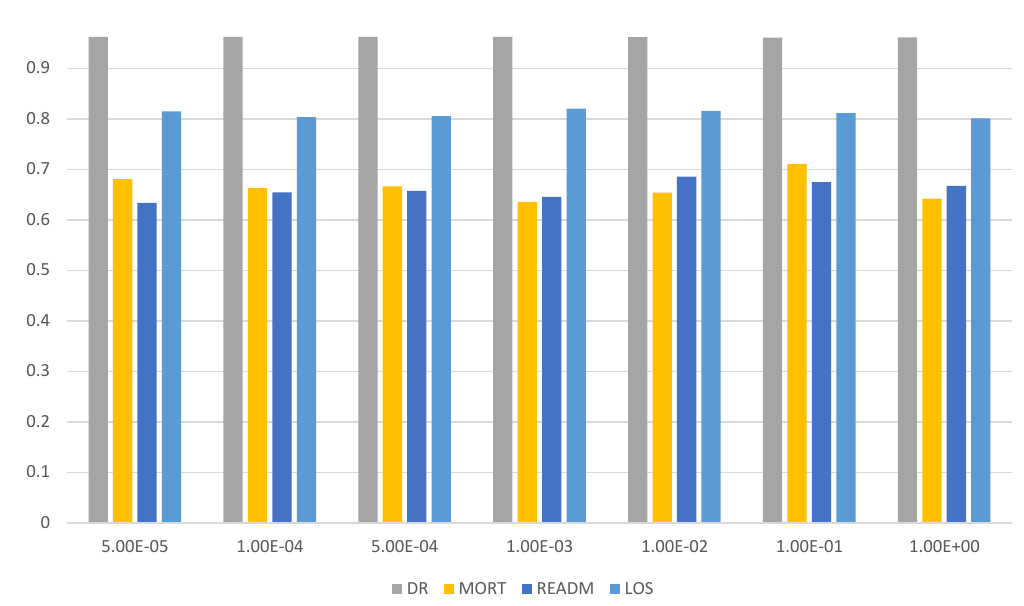}
    \caption{Performance in AUROC of MulT-EHR with different values of $\lambda$ on MIMIC-III tasks (DR: drug recommendation,
MORT: mortality, READM: readmission, LoS: length of stay).} 
    \label{fig: reg}
\end{figure}

\begin{figure}[h]
\centering
\includegraphics[width=0.45\textwidth]{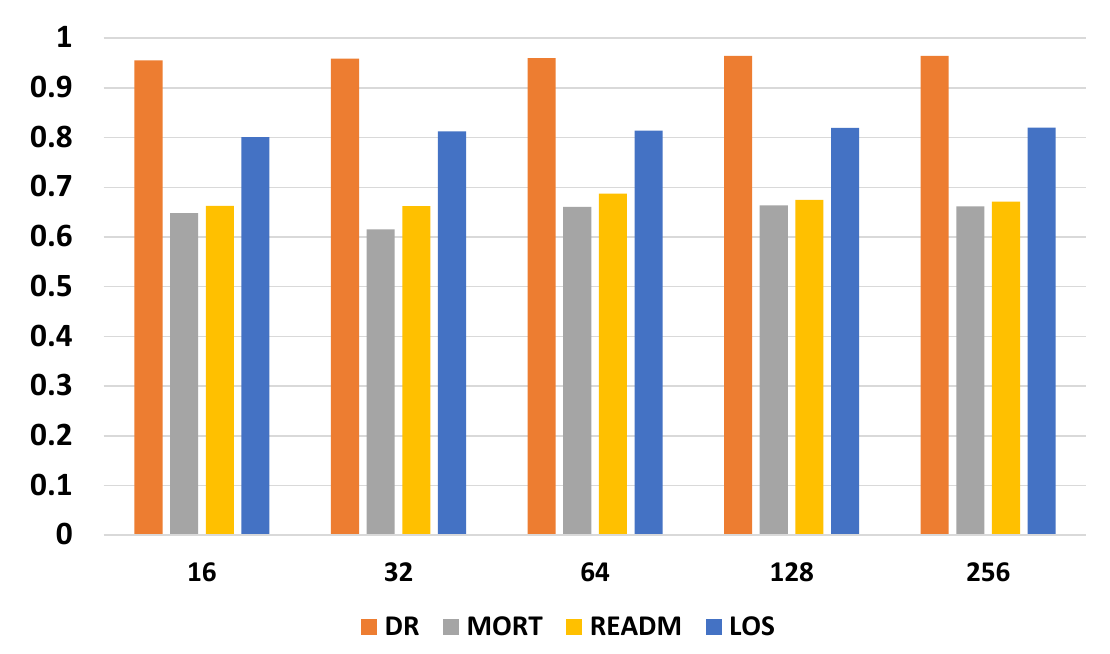}
    \caption{Performance in AUROC of MulT-EHR with different hidden dimensions on MIMIC-III tasks (DR: drug recommendation, MORT: mortality, READM: readmission, LoS: length of stay).}
    \label{fig: dimension}
\end{figure}

%--------------------------------
\subsection{Hyperparameter Tuning}
%--------------------------------

\para{Tuning Parameters for Objectives.} We evaluate the effect of the regularization parameter $\lambda$ of different task losses.
Figure \ref{fig: reg} presents the change in performance as $\lambda$ increases.
We observe that the performance is in general robust to $\lambda$, where a slightly decreased performance is observed when $\lambda$ is too large.
Since $\lambda$ represents the regularization of learning, larger $\lambda$ imposes a heavier penalty on the shortcut features and thus more restrictions on the feature space.
The predictive performance would be sacrificed as a result due to the more tightly constrained feature space

\para{Number of GNN Layers.} 
We evaluate the performance of our framework with respect to the change in the number of GNN layers, as shown in Table \ref{tab: no-layers}. 
We observe that for both datasets, as long as the GNN is not too shallow (i.e., depth $>$ 1), it can achieve satisfactory performance. 
On the other hand, the performance is slightly compromised due to
%This is likely because of 
the well-known over-smoothing problem in deeper GNNs. 

\para{Hidden Embedding Dimension.}
Figure \ref{fig: dimension} presents the comparison of different dimensions of hidden features.
We observe that the performance generally improves when a larger number of feature dimensions is adopted. 
The number of feature dimensions controls the width of the neural network. 
Hence, this verifies that increasing the width instead of depth can improve the feature representation learning performance while more effectively preventing the over-smoothing issue \citep{shi2022revisiting, zhao2019pairnorm}.

%---------------------------------------------------------
\section{Discussion and Conclusion}
%---------------------------------------------------------

%We discuss the broader impact of incorporating a heterogeneous graph model and applying causal inference in EHR analysis.
%
To address the significant confounding effects present in EHR data, we propose a denoising module based on causal inference. 
This module effectively adjusts for the confounding effects and yields causal features by eliminating most of the backdoor paths associated with trivial features. 
Not only 
do these causal features enhance predictive performance, but they also offer potential for causal explanations. While in this work we interpret the model using attention weights on causal features from the GNN model, this does not necessarily provide a causal explanation. 
Developing algorithms for causal explanations would require rigorous theoretical work to ensure causality, which falls outside of the focus of this paper.
However, based on the causal features obtained for each entity, there is a promising potential to develop a robust causal interpretation model that can identify the causes of medical events such as mortality, diseases, or readmission. 
This has significant implications for medical reasoning and future clinical research.

EHRs present a great degree of heterogeneity --- each medical entity in the EHR has a distinct node type. 
Each node type thus introduces a unique distribution of the medical entities of that type.
By using a heterogeneous graph, we can model EHRs with the awareness of different types of medical entities and relations.
Consequently, an effective heterogeneous GNN is employed to align the embedding distributions of nodes from different node types into a unified latent space. 
The transformer architecture, known for its ability to align embedding distributions across different spaces, demonstrates superior empirical performance compared to other GNN designs (as shown in Table \ref{tab: ablations_aggregation}).
Additionally, not only do EHRs include enriched medical records in the tabular form, but they also encompass information from various modalities, such as clinical/discharge reports in text and radiology scans in images. 
To incorporate this multimodal information, we merge these observations into the heterogeneous graph as individual nodes, assigning different node types based on their modalities. 
By leveraging transformer-based heterogeneous GNN architectures \citep{hu2020HGT}, we can align the distributions from different modalities, potentially leading to significant performance improvements on EHR analysis tasks.

\para{To Conclude.} We propose a novel multi-task learning framework for EHR modeling, named MulT-EHR, which mines the heterogeneity in EHR data while adjusting for confounding effects. 
Our proposed framework employs a GNN architecture that incorporates causal disentanglement for debiasing and minimizes the task-level variance to improve cross-task performance. 
Empirical studies on various datasets validate the superiority of our proposed method over existing single-task and multi-task designs. 
Qualitative analysis on node embeddings and interpretability suggests that our method can potentially provide interpretations to key medical entities (e.g., diagnosis) leading to the event (e.g., readmission).
Enriched ablation studies verify the robustness of our method to variations on the proposed components and hyperparameters. 
Our framework can potentially be generalized to any other application domains based on graph representation learning, such as recommendation systems and molecular chemistry.

\medskip

\para{Declaration of Generative AI and AI-assisted Technologies in the Writing Process.}

During the preparation of this work, the author(s) used ChatGPT in order to polish the language. After using this tool/service, the authors reviewed and edited the content as needed and take full responsibility for the content of the publication.

\para{Code and Data Availability.}

The codes for reproducing this work are available at \url{https://github.com/HKU-MedAI/MulT-EHR}.

\para{Acknowledgement.}
We thank the anonymous reviewers for their valuable comments and suggestions.
This work was partially supported by the Research Grants Council of Hong Kong (17308321 and 27206123), the Theme-based Research Scheme (T45-401/22-N), the Hong Kong Innovation and Technology Fund (ITS/273/22), and the National Natural Science Fund (62201483).

\vspace{3mm}

\bibliographystyle{plainnat}
\bibliography{references}

\appendix

%--------------------------------
\section{Evaluation Metrics}
%--------------------------------
We provide detailed definitions of the evaluation metrics.
% \begin{itemize}
% \item Classification metrics:
\begin{itemize}
    \item Accuracy: the fraction of correct predictions to the total number of ground truth labels.
    \item F-1 score: the F-1 score for each class is defined as
    \begin{align*}
        \text{F-1 score} = 2 \cdot \dfrac{\text{precision} \cdot \text{recall}}{\text{precision} + \text{recall}}
    \end{align*}
    where `recall' is the fraction of correct predictions to the total number of ground truths in each class and precision is the fraction of correct predictions to the total number of predictions in each class.  For multi-class and multi-label classification tasks, we adopt the weighted F-1 score. 
    \item AUC: the area under the receiver operating curve (ROC) which is the plot of the true positive rate (TPR/Recall) against the false positive rate (FPR).
    \item AUPR: the area under the precision-recall curve.
    \item Jaccard index: measures the similarity between the true binary labels and the predicted binary labels by the ratio of the size of the intersection of the true positive labels and the predicted positive labels to the size of their union, %of the true positive labels and the predicted positive labels,
    \begin{align*}
        \text{Jaccard} = \dfrac{\text{TP}}{\text{TP} + \text{FP} + \text{FN}}.
    \end{align*}
\end{itemize}
%\end{itemize}

%--------------------------------
\section{Additional Information on Related Methods}
%--------------------------------
In this section, we provide supplementary information on the baseline methods employed in our study. 
All baseline models were trained for 50 epochs with the option of early stopping.
\begin{itemize}
    \item RNN \citep{medsker2001RNN}: vanilla RNN predicts labels based on modeling the visit series of patients. 
    % \item LSTM \citep{hochreiter1997LSTM} \TODO{and potential variants}
    \item Transformer \citep{vaswani2017transformer}: it leverages the idea of self-attention, which allows the model to selectively focus on different parts of the input sequence when generating output. 
    \item GRAM \citep{choi2017gram}: the first work that models EHRs with a knowledge graph and uses recurrent neural networks to learn the medical code representations and predict future visit information. 
    \item StageNet \citep{gao2020stagenet}: using a stage-aware LSTM to conduct clinical predictive tasks while learning patient disease progression stage change in an unsupervised manner.
    \item Concare \citep{ma2020concare}: it considers personal characteristics during clinical visits and uses cross-head decorrelation to capture inter-dependencies among dynamic features and static baseline information for predicting patients' clinical outcomes given EHRs.
    \item Adacare \citep{ma2020adacare}: it captures the long- and short-term variations of biomarkers as clinical features, models the correlation between clinical features to enhance the ones which strongly indicate health status, and provides qualitative interpretability while maintaining the state-of-the-art performance in terms of prediction accuracy.
    \item Dr. Agent \citep{gao2020dragent}: mimics clinical second opinions using two reinforcement learning agents and learns patient embeddings with the agents.
    \item GRASP \citep{zhang2021grasp}:  GNN is used to cluster patients with their latent features and identify similar patients based on latent clusters. 
    \item GraphCare \citep{jiang2023graphcare}: it integrates external open-world knowledge graphs (KGs) into the patient-specific KGs with large language models.
\end{itemize}

\begin{itemize}
    \item Dipole \citep{ma2017dipole}: it adopts the bidirectional recurrent neural network and attention mechanism to learn medical code representations and makes prediction.
    \item KAME \citep{ma2018kame}: a generalized version of GRAM \citep{choi2017gram} by adding the attention mechanism to graph representation learning to provide interpretative diagnoses.
    \item SparcNet \cite{jing2023sparcnet}: an algorithm that can classify seizures and other seizure-like events with expert-level reliability by analyzing electroencephalograms (EEGs).
\end{itemize}

Methods particularly designed for drug recommendation tasks include:
\begin{itemize}
    \item MICRON \citep{yang2021micron}: it is a novel recurrent residual network to encode the longitudinal information of medications.
    \item Safedrug \citep{yang2021safedrug}: a drug-drug interaction controllable (DDI-controllable) drug recommendation model that leverages drugs' molecule structures and models DDIs explicitly. It uses a global message passing neural network (MPNN) module and a local bipartite learning module to fully encode the connectivity and functionality of drug molecules.
    \item MoleRec \citep{yang2023molerec}: a novel molecular substructure-aware encoding method that employs a hierarchical architecture to model inter-substructure interactions and the impact of individual substructures on a patient’s health condition.
\end{itemize}
%--------------------------------
\section{Additional Details on GNN Architectures}
%--------------------------------
We provide additional details of GNN architectures compared in the ablation studies.
We fix the number of layers as 2, and the number of hidden dimensions as 128 for different architectures. 
\begin{itemize}
    \item Graph Convolutional Network (GCN): it uses a spectral approach to define a convolution operation on the graph, and aggregates information from a node's neighbors to update the node's features.
     \item Graph Attention Network (GAT): it uses attention mechanisms to weigh the contributions of a node's neighbors. GAT uses a self-attention mechanism to calculate the attention weights and aggregates the neighbors' features based on these weights.
    \item Graph Isomorphism Network (GIN): it uses a learnable function to aggregate the features of a node's neighbors. GIN is invariant to the ordering of the neighbors and can be applied to both directed and undirected graphs.
    \item Heterogeneous Graph Transformer: it uses a multi-head attention mechanism to aggregate information from different types of nodes and edges.
    \item Relational Graph Convolutional Network: it updates node features by performing neighbour averaging by using different types of edges and different weight matrices.
\end{itemize}

\section{Additional Visualizations}

\begin{figure}
\centering
\includegraphics[width=0.5\textwidth]{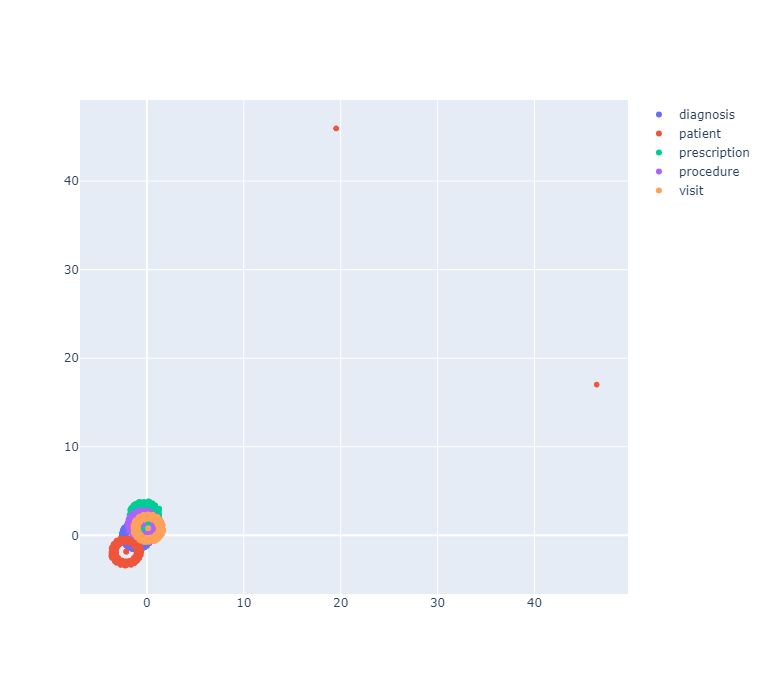}
\includegraphics[width=0.5\textwidth]{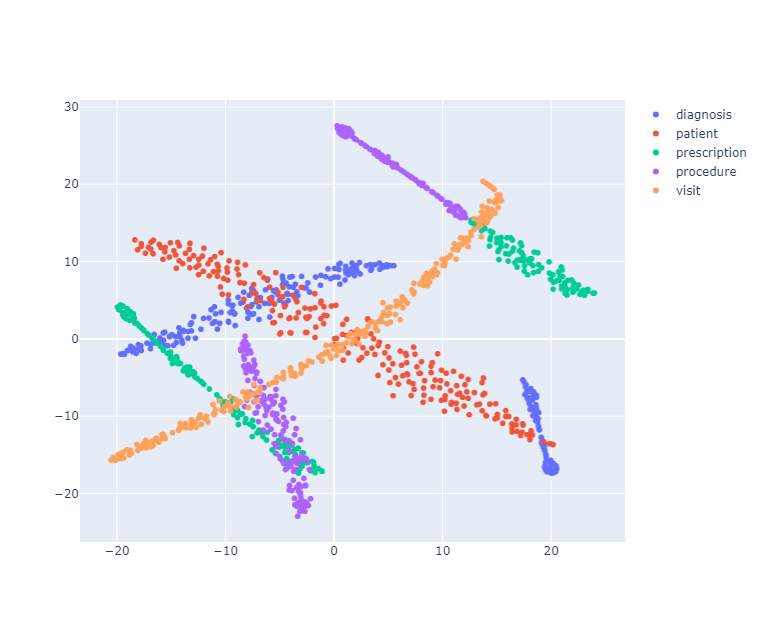}
\includegraphics[width=0.5\textwidth]{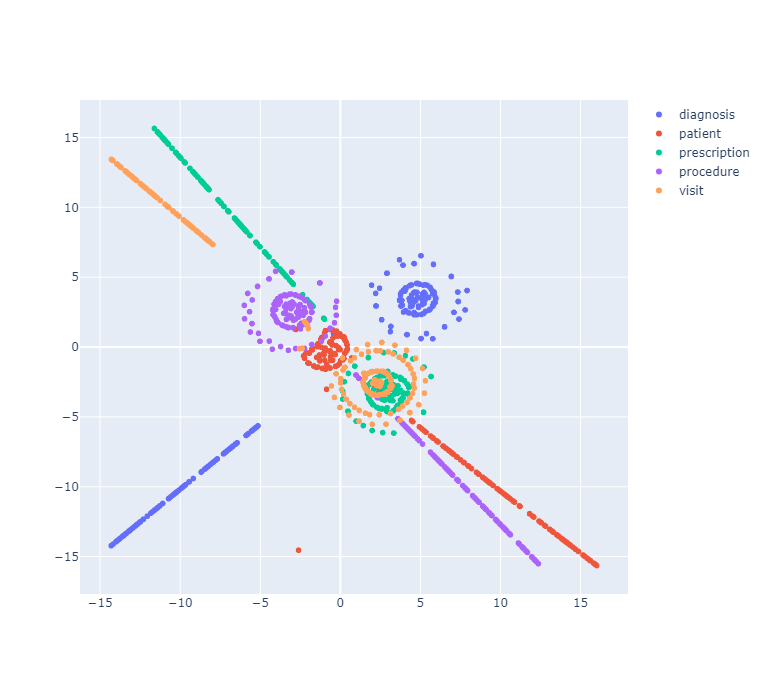}
    \caption{Node embeddings of MIMIC-III entities trained by GCN (top), GAT (middle), and GIN (bottom). }
    \label{fig: gcn_emb}
\end{figure}

\begin{figure}
\centering
\includegraphics[width=0.5\textwidth]{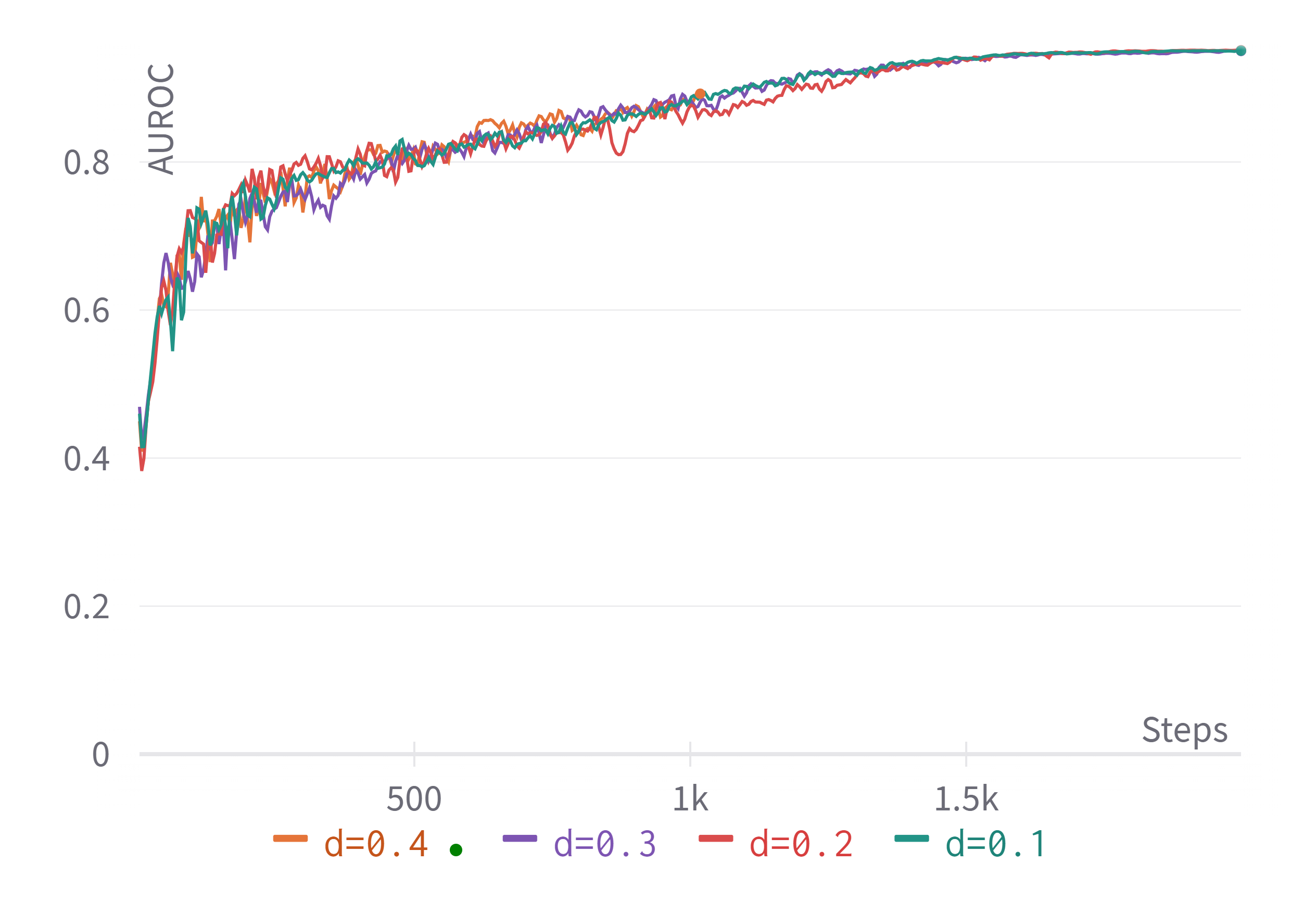}
    \caption{Learning curves of MulT-EHR with different dropout rates.}
    \label{fig: learning_curve}
\end{figure}

Figure \ref{fig: gcn_emb} visualizes the node embeddings of MIMIC-III entities trained by GCN. 
We observe that different node types form distinguishable clusters, although there are some overlaps between the clusters.
Figure \ref{fig: learning_curve} presents the learning curves of our method with respect to different dropout rates.
We observe that the dropout rates have little effect on the learning outcome. 

\section{Additional Hyperparameter Studies}

\begin{figure}
\centering
\includegraphics[width=0.45\textwidth]{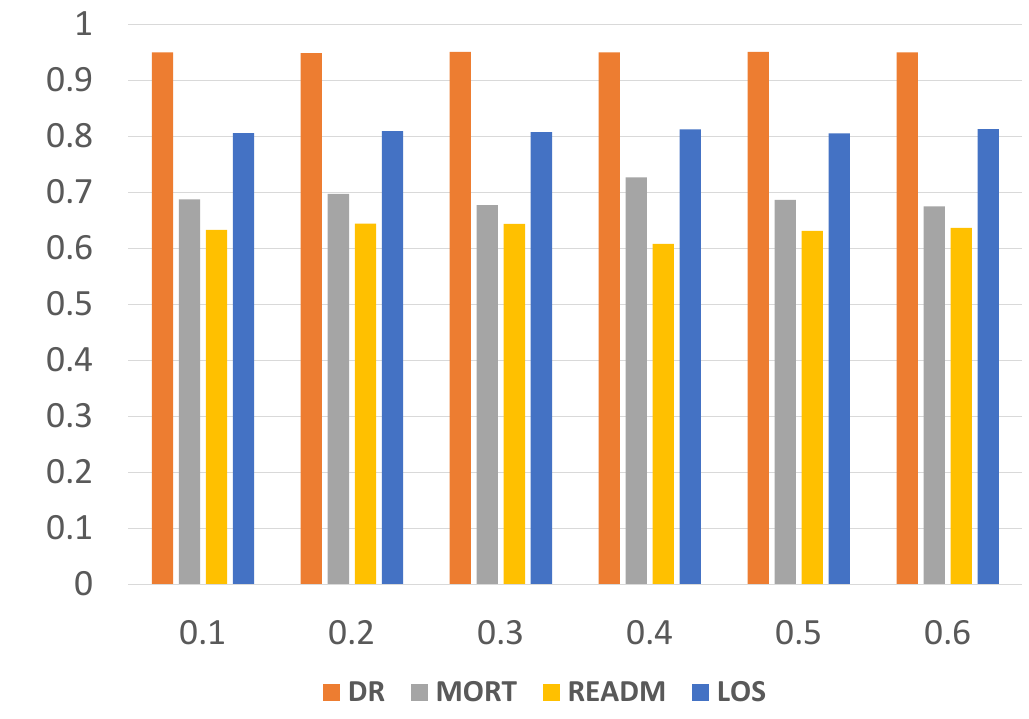}
    \caption{Performance in AUROC of MulT-EHR with different dropout rates on MIMIC-III tasks (DR: drug recommendation, MORT:
mortality, READM: readmission, LoS: length of stay).}
    \label{fig: dropout}
\end{figure}

%\para{Effects of Different dropout rates. } 
The dropout rate determines the regularization of the learning.
Figure \ref{fig: dropout} presents the results of our framework with different dropout rates.
We report the AUROC from the four benchmark tasks using dropout rates of $\{0.1, 0.2, 0.3, 0.4, 0.5, 0.6\}$.
We observe that our method is robust to different dropout rates. 
The example learning curve on drug recommendation tasks can be found in Figure \ref{fig: learning_curve}.
Incorporating dropouts can enhance the overall performance of the model, while it is essential to fine-tune the dropout rate to achieve optimal results that balance the regularization power, preventing it from being either too strong or too weak.

%\para{Settings of Other Hyperparameters.}
%
We provide the values of other hyperparameters to which our framework is not sensitive as follow:
\begin{itemize}
    \item Node feature dimension ($d$): 128
    \item Number of node samples in training: 2000
    \item Learning rate: 0.005
    \item Weight decay: 0.001
    \item Number of attention heads for attention-based algorithms: 8
\end{itemize}

\end{document}